\documentclass[10pt]{article}
\usepackage[preprint]{tmlr}

\usepackage{booktabs}
\usepackage{graphicx}
\usepackage{amsmath}
\usepackage{hyperref}
\usepackage{url}

\title{Quantifying Protocol-Induced Uncertainty in Comparative Predictive-Model Evaluation: Evidence from Large-Scale Daily PM10 Forecasting}

\author{\name Kiersten Monahan \email kiersten.monahan@eastern.edu \\
      \addr Applied Data Science program, Eastern University\\
      1300 Eagle Road, St.\ Davids, PA 19087, USA
      \AND
      \name Rafael da Silva \email rafael.dasilva@eastern.edu \\
      \addr Applied Data Science program, Eastern University\\
      1300 Eagle Road, St.\ Davids, PA 19087, USA}

\newcommand{\AggStabKRec}{3}
\newcommand{\AggStabWSBase}{35.3}
\newcommand{\AggStabFromA}{35.5}
\newcommand{\AggStabToA}{33.2}
\newcommand{\AggStabFromB}{35.5}
\newcommand{\AggStabToB}{36.7}
\newcommand{\AggStabScaleX}{7.1}
\newcommand{\AggStabNDisc}{339}
\newcommand{\AggStabEPAPlateauA}{29.0}
\newcommand{\AggStabEPAPlateauB}{31.5}

\begin{document}

\maketitle

\begin{abstract}
Comparative studies of predictive models typically end by ranking the candidates. However, the ranking depends on an evaluation protocol, and the choices that define this protocol are rarely treated as a source of uncertainty. We formalise this source as protocol-induced ranking uncertainty. Our framework applies to arbitrary candidate sets and evaluation coordinates. It compares the ranking displacement caused by switching protocols with the displacement produced by selected conventional choices within a fixed protocol. We quantify these two forms of displacement using the Protocol Sensitivity Score (PSS) and a full-refit intraprotocol null.

We provide the first empirical validation of the framework in a large-scale sequential-prediction case study of daily PM10. We compare static-split and rolling-origin evaluation across 425 European background stations and 365 US EPA monitors, analysing the two panels separately. Switching protocols yields PSS values of 0.801 and 0.772 and changes the selected model at 35.3\% and 31.5\% of stations. In Europe, intraprotocol perturbations with identical scored targets yield PSS values of 0.072 and 0.230, with winner-swap rates of 0.8\% and 4.9\%. These values are three to seven times smaller than the between-protocol discrepancy. Expanding the candidate set from three to nine models raises the between-protocol winner-swap rate to 60.2\%. The pattern remains under a frozen protocol applied to held-out background and non-background stations. We recommend reporting ranking variance under a small set of legitimate intraprotocol perturbations alongside claims of model superiority.
\end{abstract}

\section{Introduction}
\label{sec:intro}

A comparative evaluation of predictive models usually ends with a ranking: model A outperforms B, which outperforms C. This ranking often guides practical decisions, including which model to deploy, which architecture to develop, and which research direction to pursue. However, the ranking depends on an evaluation protocol. That protocol includes choices that are often under-reported, such as how to separate fitting from evaluation, how often to refit the models, and which time window to score. We therefore ask: \textbf{how much does a published ranking change across legitimate evaluation protocols, relative to the displacement observed when the protocol is held fixed?}

This question differs from asking which protocol best estimates generalisation error. Empirical machine learning has long examined how to compare algorithms statistically \citep{dietterich1998,demsar2006}. More recent work has shown that benchmark conclusions are sensitive to several sources of evaluation variance \citep{bouthillier2021,pineau2021}. This literature mainly asks an \textbf{estimation} or \textbf{significance} question: is a reported performance difference trustworthy under a fixed evaluation protocol? Our study asks a \textbf{decision} question: does the choice of protocol change which model would be selected?

These questions are not equivalent. A protocol can estimate errors with bias and still preserve the correct ranking. Conversely, two legitimate protocols can produce reasonable error estimates but select different winners. In sequential prediction, this issue appears in choices such as static chronological split and rolling-origin evaluation \citep{tashman2000,bergmeir2012,cerqueira2020,hewamalage2023}. The broader methodological gap is that \textbf{comparative evaluation studies rarely quantify the uncertainty introduced by protocol selection relative to the noise that remains when the protocol is held fixed.}

The study was motivated by a ranking reversal reported for three forecasting models at a single PM10 monitoring station. The selected model changed when the evaluation protocol changed \citep{garciacrespi2026}. This observation raised a broader empirical question: was the reversal specific to one series, or did it indicate systematic sensitivity of model selection to the validation protocol? We examine this question at panel scale and formalise the corresponding evaluation problem.

Answering the question requires three elements. First, we need a continuous measure of ranking displacement, rather than only a count of inversions, so that the magnitude can be compared across evaluation units. Second, we need panel-scale evidence with quantified uncertainty, rather than evidence from a single site. Third, we need a \textbf{null reference}: an estimate of how much displacement occurs when the protocol is held fixed and only a conventional choice within that protocol is changed. Without this reference, a large difference between protocols cannot be interpreted relative to the instability that remains within a fixed protocol.

To the best of our knowledge, prior work addresses parts of this problem but not the full comparison. For example, one study examines ranking instability across protocol dimensions using a permutation null in another domain \citep{kendiukhov2026}, while another examines the effect of protocol choice on error estimates in sequential settings \citep{cerqueira2020}. Neither provides a panel-scale, reference-based comparison between protocol-induced displacement and intraprotocol displacement under full model refitting.

We implement this reference-based attribution in a sequential predictive-evaluation case study. At each monitoring site, we evaluate three models under two protocols and across seven forecast horizons. The Protocol Sensitivity Score (PSS) measures displacement of the complete ranking. A second measure, winner swap, records whether the protocol changes the selected model.

We then construct an intraprotocol reference for each protocol and fully refit the models after every change. For the static split, we perturb the training boundary. For rolling-origin evaluation, we perturb the refit calendar. In both cases, the compared runs score exactly the same targets. This condition is essential: as we later show, changing the scored targets can reverse the attribution result. The analysis therefore compares the observed between-protocol discrepancy with these two intraprotocol references.

The study produces three main findings. First, the discrepancy between protocols is three to seven times larger than the internal noise measured within either protocol. We also test two alternative explanations---unequal test sets and training-data lag---and the data do not support them as the main source of the discrepancy.

Second, the smaller intraprotocol displacement is still meaningful. Rankings change perceptibly between two legitimate runs of the same protocol when the refit calendar is shifted by only two weeks. This is a conventional choice that comparative studies rarely report.

Third, in the candidate-set expansion examined here, increasing the number of models from three to nine raises the between-protocol winner-swap rate from 35.3\% to 60.2\%. A larger decision space therefore does not stabilise model selection in this case study.

The case study evaluates daily PM10 prediction in two independent monitoring networks. The confirmatory Heisig/EEA panel contains 425 European background stations drawn from 1216 eligible stations. The transfer panel contains 365 US EPA AirData monitors. We analyse the panels separately rather than pooling them because they play different inferential roles: the European panel is confirmatory, whereas the US panel tests transfer to another network.

We conduct two additional checks. First, we apply a frozen protocol once to 271 background stations that were not previously examined and to 520 traffic and industrial stations. Second, we repeat the full analysis with nine models and re-derive the equivalence margin for the larger candidate set. The validated claims remain limited to this sequential daily-PM10 setting. Section~\ref{sec:limitations} states these limits, and Section~\ref{sec:future} describes the corresponding research programme.

The research questions apply the general quantification problem to this case study:
\begin{enumerate}
\item \textbf{RQ1.} How large is the ranking displacement between static-split and rolling-origin evaluation, how is it distributed across stations, and how often does it change the selected model?
\item \textbf{RQ2.} How does the displacement between protocols compare with the displacement produced by the selected conventions within each protocol?
\item \textbf{RQ3.} Does the reference-based attribution remain supported when we expand the candidate set, analyse previously unexamined stations, include site strata excluded from the original design, and repeat the analysis in a network on another continent?
\item \textbf{RQ4.} Can observable features of the series predict where instability will be larger with enough accuracy to support a diagnostic index?
\end{enumerate}

These questions correspond to five hypotheses:
\begin{enumerate}
\item[\textbf{H1.}] Between-protocol displacement is not equivalent to zero under margin $\delta$.
\item[\textbf{H2.}] Between-protocol displacement exceeds the displacement produced by the selected intraprotocol perturbations.
\item[\textbf{H3.}] The reference-based decomposition remains supported with nine models.
\item[\textbf{H4.}] The pattern reappears in reserved material, in non-background strata, and in the EPA network.
\item[\textbf{H5.}] Observable series features predict PSS above the pre-specified viability threshold.
\end{enumerate}

The remainder of the paper is organised as follows. Section~\ref{sec:related} positions the contribution within the literature on empirical evaluation. Section~\ref{sec:data} describes the data and eligibility criteria. Section~\ref{sec:methods} presents the study design and statistical plan, and Section~\ref{sec:results} reports the results. Section~\ref{sec:discussion} discusses attempted refutations, limitations, and implications for comparative predictive-model evaluation.

\section{Related work}
\label{sec:related}

\textbf{Statistical comparison of algorithms.} Machine learning has a long tradition of testing whether one learning algorithm outperforms another under a fixed evaluation design \citep{dietterich1998,demsar2006}. This literature addresses Type~I error control and comparisons involving multiple algorithms and datasets after a protocol has been selected. Our study varies the protocol itself. We ask how much ranking disagreement arises when two legitimate protocols are applied in practice, and how much of that disagreement can be attributed to the protocol contrast.

\textbf{Evaluation variance and reproducibility.} Recent empirical ML shows that published comparisons can be sensitive to data sampling, model initialisation, and hyperparameter choices. Reproducible evaluation therefore requires these sources of variation to be made explicit \citep{bouthillier2021,pineau2021}. Related work also shows that faults in an evaluation pipeline, such as data leakage, can inflate apparent performance gains \citep{kapoor2023}.

This literature primarily examines the fragility of reported scores within an evaluation procedure or the effects of pipeline faults. We examine a different source of instability: the selection of the evaluation protocol itself. More specifically, we compare ranking displacement caused by \textbf{protocol selection} with displacement caused by a selected \textbf{intraprotocol convention}. The object of interest is the ranking among models, not the score of a single model.

\textbf{Variance of evaluation estimators.} Closely related studies examine both the bias and the variance of resampling-based performance estimators. They also show how estimator variance can make algorithm comparisons fragile \citep{braganeto2004,kim2009,molinaro2005}. Their primary object is the error estimate of a model. Our primary object is the resulting decision: the ranking among the candidate models.

We therefore measure how much the ranking moves between two legitimate runs of the same protocol when only a conventional choice is changed. This intraprotocol displacement provides an empirical reference for interpreting the larger discrepancy observed between protocols.

\textbf{Complementary critiques of evaluation protocols.} TabArena \citep{erickson2025} and TabReD \citep{rubachev2024} show that evaluation conditions can affect published rankings in tabular learning. These settings do not contain the sequential temporal structure of our case study.

\citet{kendiukhov2026} quantify ranking instability across evaluation-protocol dimensions in gene-regulatory-network benchmarking. They use pairwise reversal rates and a permutation null, and find that the observed instability is well below what would be expected from random rankings. Their design shares our premise that published rankings can be protocol-sensitive, but the two approaches differ in three ways. First, their null permutes ranks, whereas our reference changes a legitimate convention within a protocol and fully refits the models. Second, our compared runs use identical scored targets. Third, our analysis explicitly separates displacement between protocols from displacement within a fixed protocol in a sequential setting.

A different critique concerns benchmark composition. \citet{saqur2026} argue that gains in time-series forecasting can appear larger when benchmark suites are dominated by series with strong periodicity. Their mechanism is suite selection rather than data partitioning or validation, and their target is illusory accuracy gains rather than ranking displacement across protocols.

Together, these studies show that comparative conclusions in predictive modelling can depend on evaluation choices that are rarely reported. The studies differ in which choice is varied and in the mechanism through which that choice affects the conclusion.

\textbf{Sequential prediction protocols (case-domain literature).} In temporally ordered prediction, validation design is known to affect estimated error. Out-of-sample testing and rolling-origin evaluation are established options \citep{tashman2000,bergmeir2012,cerqueira2020,hewamalage2023}. This literature mainly asks which protocol estimates prediction error more reliably. We instead ask how often two protocols used in practice select different models. We answer this question at panel scale and interpret the difference against an intraprotocol reference.

Research on multi-step air-quality prediction has focused mainly on model architecture and predictive accuracy \citep{cabaneros2019,abuouelezz2025}. \citet{garciacrespi2026} report that changing the protocol reversed the ranking of models at one station. This result provides the empirical starting point for our study. We test whether ranking displacement also appears across large station panels and whether it exceeds the variability observed within a fixed protocol.

The panel-level claims do not require the original single-site result to be accepted as valid. We analyse the corresponding re-execution separately in Section~\ref{sec:singlesite} and Supplementary Section~SM-08.

\textbf{Contributions to empirical evaluation methodology.} To the best of our knowledge, these lines of work do not provide a panel-scale comparison of ranking displacement between protocols with intraprotocol reference perturbations based on full model refitting. We propose four elements and validate them in one sequential-prediction case study:
\begin{enumerate}
\item A general formulation of the Protocol Sensitivity Score for arbitrary candidate sets and evaluation coordinates. We pair PSS with winner swap so that displacement of the full ranking remains distinct from a change in the selected model.
\item An intraprotocol reference constructed with full refitting and identical scored targets. This reference makes the magnitude of between-protocol displacement interpretable relative to sample change and variation within the procedure.
\item A structured set of robustness checks and tests of alternative explanations. Each check has an executable analysis and an interpretation rule. This set includes a diagnostic index that did not pass its pre-set viability threshold (adj.\ $R^2=0.105<0.15$; decision \texttt{SUPPLEMENTARY}).
\item An analysis of how \textbf{decision-space size} relates to selection stability. In the candidate-set expansion examined here, adding models increases protocol-dependent changes in model selection.
\end{enumerate}

\section{Data}
\label{sec:data}

The design uses two independent monitoring networks that measure the same pollutant. We apply the same eligibility criterion to both networks so that differences in data curation do not define the cross-continental comparison. This section describes the networks, the common eligibility criterion, and the selection funnel from the available universe to each analysed panel. The eligibility criteria were fixed before analysis (Supplementary Section~SM-02).

\subsection{Confirmatory European panel (Heisig / EEA)}
\label{sec:eea}

The primary source is the snapshot published by \citet{heisig2024} on Zenodo under DOI. It contains daily PM10 observations from 2015 to 2023 and station metadata, including coordinates, country, and site type.

The choice between a live EEA service and a frozen archive directly affects reproducibility. We use the frozen archive. The snapshot is a fixed package, so a replication that downloads the same DOI receives the same bytes.

A live source can change between runs because of retrospective corrections or because stations are added or removed. In that setting, a difference between results could reflect either the analysis or an unrecorded change in the input data. Ending the series in 2023 is therefore a deliberate design choice. The audit evaluates a static and verifiable dataset rather than the most recent available data.

We report the full provenance chain: the EEA is the original data producer, and the Heisig snapshot is the versioned compilation used in the analysis \citep{heisig2024}. The confirmatory European panel is drawn \textbf{exclusively} from this snapshot.

We also inventoried a second EEA compilation, the provisional airbase set. It contains 2615 stations, of which 1172 meet the same eligibility criterion. This set remains outside confirmatory inference and is not combined with the Heisig panel. Combining the compilations would place stations with different curation and verification status under one aggregated $N$.

The European selection funnel has three stages. The Heisig universe contains 2807 stations with PM10 data. After applying the record-length and completeness criteria, 1216 stations remain eligible. Of these, 425 belong to the \textbf{background} stratum and form the confirmatory panel.

Restricting the confirmatory panel to background stations reduces the risk of conflating protocol sensitivity with local emission effects. It also aligns the panel with the station examined in the complementary module. The 425 stations are not a convenience or computational subsample; they are the \textbf{census} of eligible background stations. We apply the complete protocol---two evaluation protocols, three models, and seven horizons---to every station in this census.

Traffic and industrial stations are excluded from confirmatory PSS inference for the background stratum. They are used later for frozen-protocol confirmation (Section~\ref{sec:holdout}) and for a secondary exceedance analysis (Supplementary Section~SM-07).

\subsection{External panel --- US EPA AirData (H4)}
\label{sec:epa}

The transfer test uses the US EPA AirData/AQS network to add geographic breadth. Before analysis, we checked its selection as the external network against the pre-specified scope criterion. We use daily PM10 observations for parameter 81102 from 2015 to 2023 \citep{epa_airdata,epa_aqs}.

We apply the \textbf{same} eligibility criterion used in Europe: at least 5.0 years of data and completeness of at least 0.8. The resulting census contains 365 eligible monitors. This exceeds the pre-specified minimum of 50 monitors for an external network.

We run the same confirmatory protocol on the EPA panel but analyse it separately from the European panel. We do not pool the two networks into a single $N$. This separation is necessary for the transfer question to remain interpretable. The relevant question is whether the phenomenon reappears in another network, not what the average is across 790 stations with different provenance.

The diagnostic index did not pass its viability gate (Section~\ref{sec:h5}; Supplementary Section~SM-06). The EPA panel therefore tests transfer of the \emph{phenomenon}; it is not used to score or validate that index.

\subsection{Eligibility, sampling, and panel summary}
\label{sec:eligibility}

Both networks use the same eligibility rules: a minimum record length of 5.0 years, minimum completeness of 0.8, and an explicit interpolation rule for short gaps. The pre-specified minimum for the European network was 250 eligible stations; 1216 stations meet the criteria. The minimum for an external network was 50 monitors; 365 EPA monitors meet the criteria. Supplementary Section~SM-09 reports the eligibility census.

For each network, the sampling frame is the full eligible universe rather than a convenience subsample. Consequently, the prevalences reported in Section~\ref{sec:results} describe all stations in the corresponding network that meet the eligibility criteria. They do not describe a subset selected by the authors.

\begin{table}[t]
\centering
\caption{Panel summary for H1 and H4 claims.}
\label{tab:panels}
\begin{tabular}{@{}llp{1.4cm}p{1.2cm}p{1.2cm}ccp{2.2cm}@{}}
\toprule
Panel & Network & Universe & Eligible & $N$ & Blocks & Role \\
\midrule
EEA Heisig background & Heisig & 2807 & 1216 & 425 & 15 & confirmatory core \\
EPA AirData & EPA & -- & 365 & 365 & 38 & cross-continental \\
EEA airbase (excluded) & airbase & 2615 & 1172 & -- & -- & outside confirmatory $N$ \\
Single-site (Elche) & EEA local & 1 & 1 & 1 & -- & not aggregated \\
\bottomrule
\end{tabular}
\end{table}

\begin{figure}[htbp]
\centering
\includegraphics[width=0.80\linewidth]{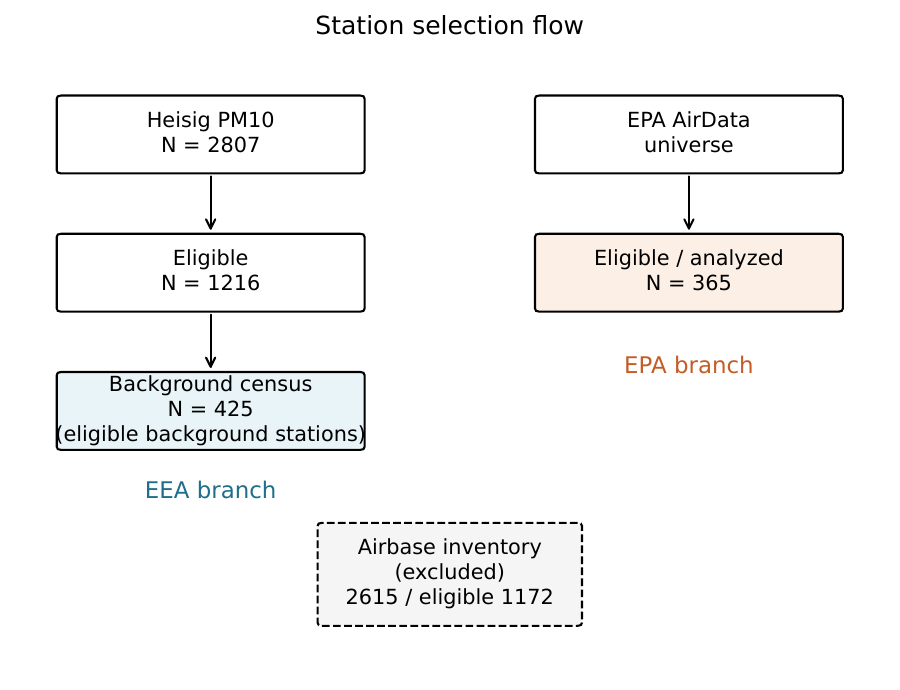}
\caption{Station selection funnel by network---universe, eligibility, European background stratum, and US census---with branches kept separate (no pooling).}
\label{fig:funnel}
\end{figure}

Table~\ref{tab:panels} summarises the panels used to evaluate H1 and H4. Figure~\ref{fig:funnel} shows the selection funnel for each network, including the available universe, eligibility stage, European background stratum, and US census. The branches remain separate because the panels are not pooled.

\subsection{Material reserved for confirmation}
\label{sec:reserved}

Among the 1216 Heisig-eligible stations, the 425 background stations described above form the confirmatory panel. The other 791 stations were not used in that inference and are reserved for confirmation. This material contains 271 additional background stations and 520 traffic or industrial stations. Section~\ref{sec:methods} describes their use, and Section~\ref{sec:holdout} reports the results.

The EEA/airbase compilation contains 2615 stations, of which 1172 are eligible. It remains entirely outside the confirmatory and confirmation analyses for the provenance reasons stated in Section~\ref{sec:eea}.

\subsection{Mechanism-module station}
\label{sec:elche}

The complementary single-site module uses the EEA version of the Elche urban-background station examined by \citet{garciacrespi2026}. We analyse this module separately and never aggregate it with either panel. Supplementary Section~SM-08 provides the details.

Licence and ethics statements are provided in Supplementary Section~SM-01.

\section{Methods}
\label{sec:methods}

\begin{figure}[htbp]
\centering
\includegraphics[width=1.0\linewidth]{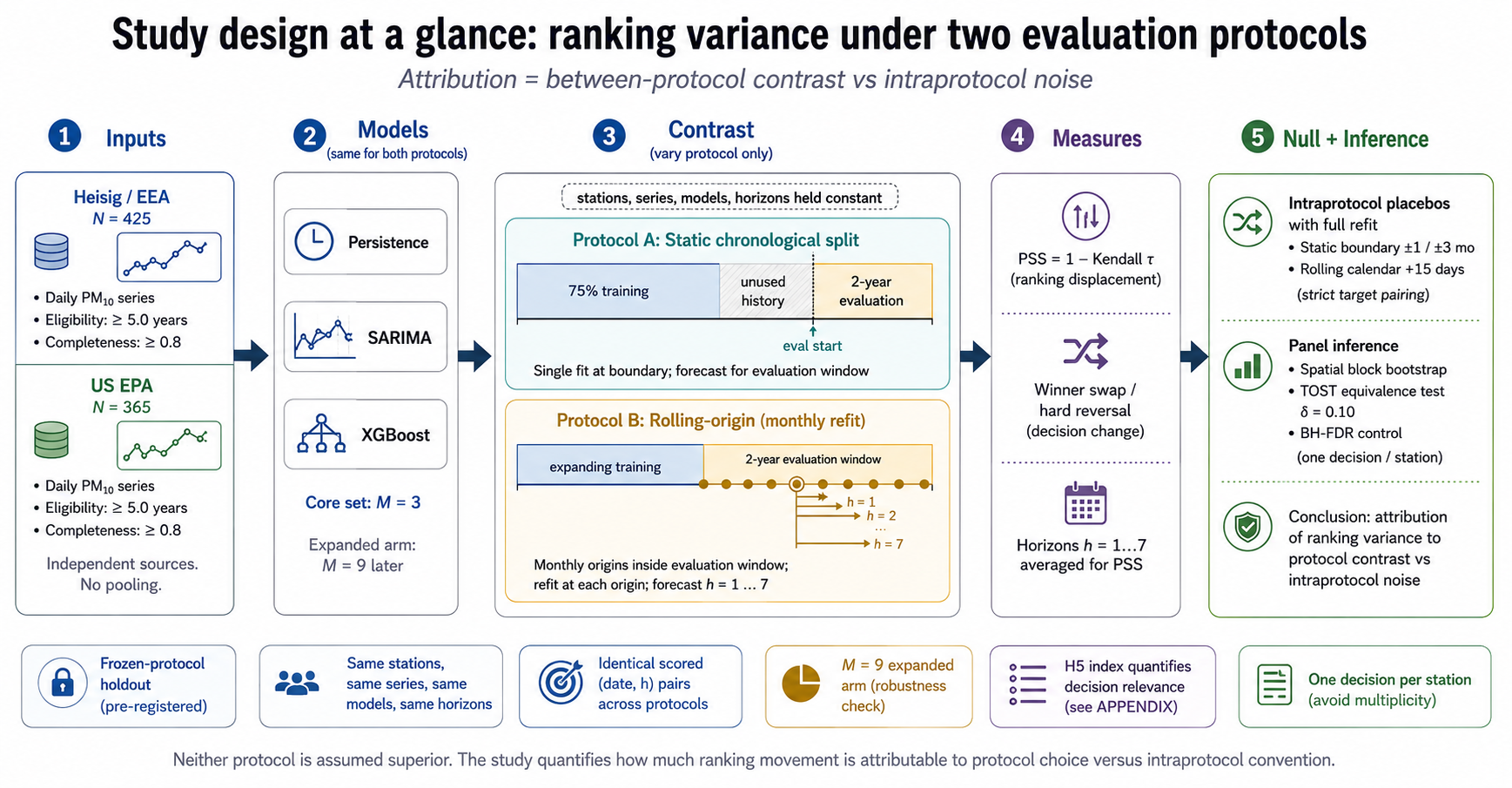}
\caption{Study design at a glance: ranking variance under two evaluation protocols. Attribution contrasts between-protocol discrepancy with intraprotocol noise.}
\label{fig:study_design}
\end{figure}

The design addresses one attribution question. With stations, series, models, and horizons held fixed, how much does the ranking change when only the validation protocol changes? We first define the two protocols and the measures used to compare their rankings. We then construct an intraprotocol reference. This reference shows how much displacement remains when the protocol is held fixed and only a convention within it changes. Finally, we describe the panel-level inference plan and the pre-commitments that support the confirmatory interpretation. Supplementary Section~SM-02 reports the protocol chronology. Figure~\ref{fig:study_design} summarises the design.

\subsection{Study architecture}
\label{sec:pipeline}

The analyses follow a fixed sequence:
\begin{enumerate}
\item determine eligibility and freeze the data;
\item conduct the pre-panel instrument check;
\item evaluate the paired protocols;
\item conduct the confirmatory analysis on the European panel (H1);
\item construct the intraprotocol reference (H2);
\item expand the candidate set from three to nine models (H3); and
\item test transfer on the EPA panel and on reserved European material (H4).
\end{enumerate}
The diagnostic-index, exceedance, and single-site branches are secondary. They do not support the main claims. The European and US panels remain separate throughout the analysis. Supplementary Section~SM-02 reports the complete execution graph, decision gates, and branch chronology.

\subsection{Models}
\label{sec:models}

The primary model set is intentionally small and classical. It contains \textbf{persistence}, \textbf{SARIMA}, and \textbf{XGBoost}. These models represent three families of increasing complexity: naive temporal inertia, parametric stochastic structure, and nonlinear learning.

The same composition is used in the illustrative Elche case \citep{garciacrespi2026} (Supplementary Section~SM-08), which preserves comparability with that examination. Persistence provides the required naive baseline. Without it, temporal inertia could be mistaken for predictive skill.

More recent architectures are excluded from the primary arm by design. The object of this arm is sensitivity of the ranking to the protocol. A set of three models is sufficient for a ranking to exist and to reverse. Adding more candidates at this stage would increase ranking variance and make Kendall's $\tau$ harder to interpret. We examine a larger candidate set separately under H3.

\subsection{General formulation of protocol-induced ranking uncertainty}
\label{sec:general}

Let $\mathcal{U}$ denote the set of evaluation units. Depending on the application, a unit may be a dataset, site, or task. Let $\mathcal{M}=\{m_1,\ldots,m_M\}$ be the common set of candidate models. For each unit $u$, let $\mathcal{C}_u$ denote its evaluation coordinates, such as horizons, folds, subgroups, or metrics.

An evaluation protocol is denoted by $p$. A conventional setting within that protocol is denoted by $\gamma$. Together, $(p,\gamma)$ produces a loss $\ell_{u,c}^{(p,\gamma)}(m)$ for candidate $m$ at coordinate $c$ of unit $u$. These losses are computed on a pre-specified set of scored targets and induce a ranking $r_{u,c}^{(p,\gamma)}$ of the $M$ candidates. We assume that lower loss is better. Metrics with the opposite orientation are transformed before ranking.

We compare two protocol configurations, $(p,\gamma)$ and $(q,\eta)$, on identical scored targets. At each evaluation coordinate, we measure the difference between their rankings as $1-\tau$, where $\tau$ is Kendall's rank-correlation coefficient \citep{kendall1938}. We then average this displacement across the coordinates. The resulting unit-level Protocol Sensitivity Score is
\begin{equation}
\mathrm{PSS}_{u}\!\left[(p,\gamma),(q,\eta)\right]
= \sum_{c\in\mathcal{C}_u} w_{u,c}\left[1-\tau\!\left(r_{u,c}^{(p,\gamma)},r_{u,c}^{(q,\eta)}\right)\right],
\label{eq:pss-general}
\end{equation}
where $w_{u,c}\ge 0$ and $\sum_c w_{u,c}=1$.

The score lies in $[0,2]$. A value of zero denotes identical rankings, one denotes no ordinal concordance, and two denotes complete reversal. In this study, the horizons receive equal weights and all rankings are complete and have no ties. Applications with tied performance should pre-specify a tie-aware coefficient, such as Kendall's $\tau_b$, rather than break ties after observing the results.

PSS describes movement in the complete ranking. We also need a decision-level measure that records whether the selected model changes. At each coordinate, we compare the candidates with the lowest loss under the two configurations. Winner swap equals one when these selected candidates differ on a majority of the weighted coordinates:
\begin{equation}
\mathrm{WS}_{u}\!\left[(p,\gamma),(q,\eta)\right]
= \mathbf{1}\!\left\{\sum_{c} w_{u,c}\,\mathbf{1}\!\left[\operatorname*{arg\,min}_{m}\ell_{u,c}^{(p,\gamma)}(m) \neq \operatorname*{arg\,min}_{m}\ell_{u,c}^{(q,\eta)}(m)\right] > \tfrac{1}{2}\right\},
\label{eq:winner-swap}
\end{equation}
Thus, PSS measures displacement of the complete ranking, whereas winner swap measures the Top-1 consequence.

PSS has a discrete resolution when rankings are complete and have no ties. One pair inversion changes $1-\tau$ by $4/[M(M-1)]$. With $H$ equally weighted coordinates, the smallest non-zero average displacement is
\begin{equation}
\delta_{\min}(M,H)=\frac{4}{M(M-1)H},
\label{eq:grid-step}
\end{equation}
For $M=3$ and $H=7$, this value is $2/21\approx0.095$. For $M=9$ and $H=7$, it is $0.0079$. This resolution rule determines the equivalence margins used below.

We next separate displacement caused by switching protocols from displacement caused by conventions within a fixed protocol. Let $\gamma_0$ and $\eta_0$ denote the reference settings for protocols $p$ and $q$. The between-protocol displacement is
\begin{equation}
B_u(p,q)=\mathrm{PSS}_{u}\!\left[(p,\gamma_0),(q,\eta_0)\right],
\end{equation}
For protocol $p$, let $\gamma_a$ and $\gamma_b$ be two legitimate settings in $\Gamma_p$. Their within-protocol displacement is
\begin{equation}
N_{u,p}(\gamma_a,\gamma_b)=\mathrm{PSS}_{u}\!\left[(p,\gamma_a),(p,\gamma_b)\right].
\end{equation}

At panel level, attribution compares the distribution of $B_u$ with the corresponding distributions of $N_{u,p}$. We use paired differences, uncertainty intervals, and descriptive ratios when the denominators are positive.

This comparison is an \emph{operational empirical decomposition}; it is not an additive ANOVA identity. Here, ``ranking variance'' refers to the distribution of ranking displacement produced by legitimate changes in protocols and conventions. It does not refer to the numerical variance of rank labels. The attribution is also relative to the pre-specified intraprotocol perturbations examined in this study. It is not an exhaustive decomposition over every legitimate convention within either protocol.

The formulation is general, but this study provides its first empirical instantiation. The validated claims are therefore limited to the sequential case study described below.

\subsection{Case-study protocols and metric instantiation}
\label{sec:protocols}

In the case study, monitoring sites are the evaluation units. Forecast horizons $h=1,\ldots,7$ are the equally weighted evaluation coordinates. Every site uses the same candidate set and supplies the same data to two protocols:
\begin{enumerate}
\item \textbf{Static split.} The series is divided once in chronological order. Training ends at the valid observation corresponding to 75\% of the history before the evaluation window.

This specification has a structural consequence that is not the intended contrast. The final quarter of the history before evaluation is not used to fit the static model. At the median station, the fit therefore ends about 637 days before evaluation begins in the European panel and 625 days before it begins in the US panel. In contrast, rolling-origin evaluation refits the models up to each origin.

We retain the original static-split specification to preserve comparability with the single-site case that motivated the study. We also analyse a contiguous-training variant to test whether training recency primarily explains the observed discrepancy. Section~\ref{sec:limitations} reports that analysis.
\item \textbf{Rolling-origin.} The evaluation uses a moving origin, monthly model refitting, and forecast horizons $h=1,\ldots,7$ days.
\end{enumerate}

Both protocols use the last two years of each series as the evaluation window. They also use the same scored origins: the start of each month. The number of evaluated points per station is therefore identical under the two protocols in both panels. Consequently, differences in test-set size cannot explain the observed ranking discrepancy.

\begin{figure}[htbp]
\centering
\includegraphics[width=1.0\linewidth]{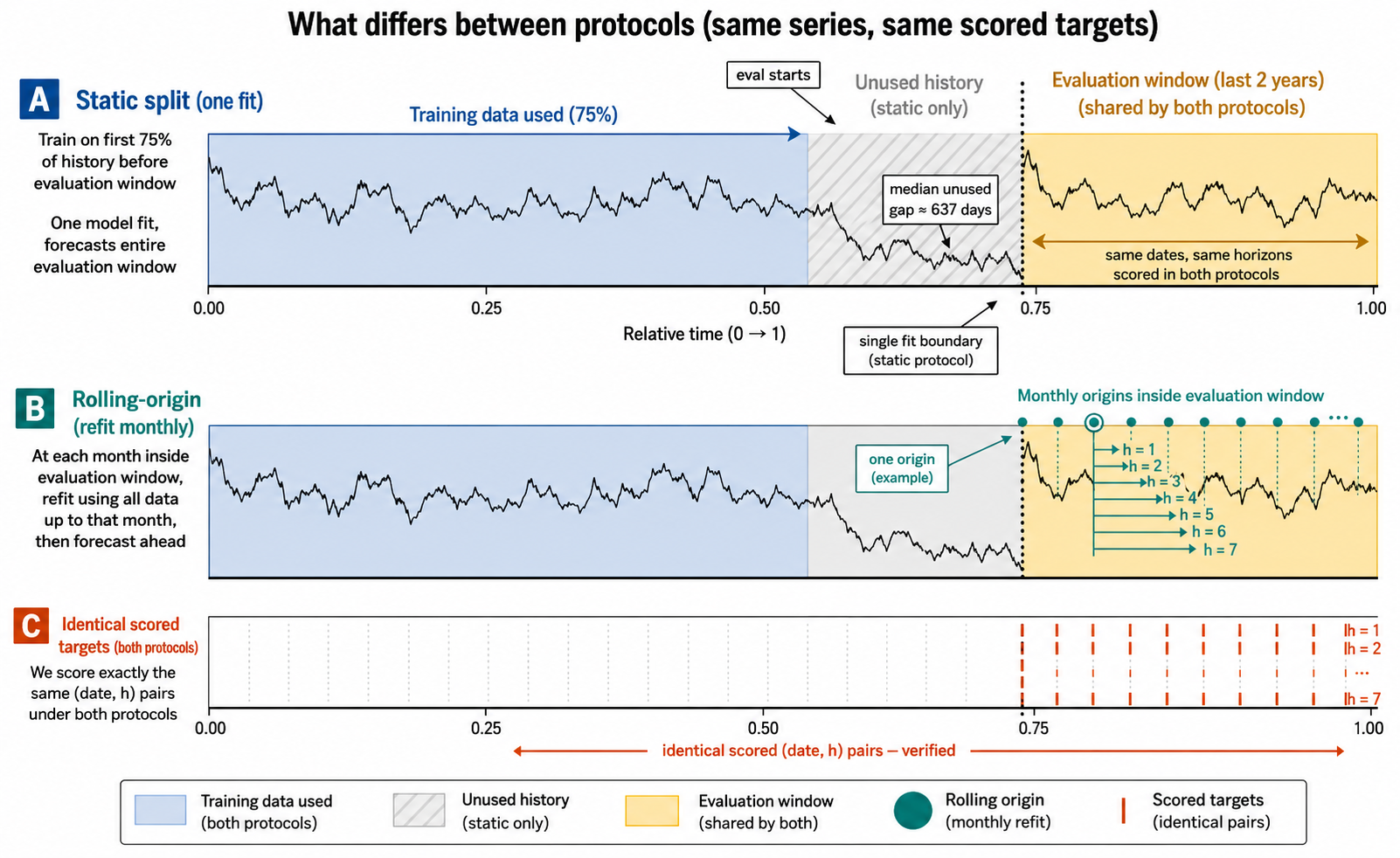}
\caption{Static-split versus rolling-origin protocols on a shared timeline: both use the full series; evaluation origins differ.}
\label{fig:protocols}
\end{figure}

Figure~\ref{fig:protocols} places the two protocols on a common timeline. It shows the point, or sequence of points, at which evaluation begins under each protocol.

\textbf{Primary and decision measures.} We evaluate Equation~\ref{eq:pss-general} with equal weights across the seven horizons. PSS is therefore the mean of $1-\tau$ across the seven horizon-specific rankings. In the three-model arm, the grid step is $2/21\approx0.095$. The nine-model arm uses the corresponding value from Equation~\ref{eq:grid-step}.

In the case study, we call the event in Equation~\ref{eq:winner-swap} a hard reversal. It occurs when the static-split winner is not the rolling-origin winner on at least four of the seven horizons. We also computed an alternative severity criterion based on exclusion from the Model Confidence Set. However, this criterion is practically inert: it occurs at only 0.24\% of stations and does not materially contribute to the reported prevalence. Supplementary Section~SM-03 reports this analysis and comparisons with Spearman correlation and rank-biased overlap.

\subsection{Intraprotocol reference and attribution rule}
\label{sec:null}

Between-protocol PSS is interpretable only relative to the displacement that remains when the protocol does not change. We estimate this reference, $N_{u,p}$, by changing one legitimate convention within a protocol and fully refitting every model.

For the static split, we move the training boundary by $\pm1$ and $\pm3$ months. For rolling-origin evaluation, we move the refit calendar by fifteen days. These are plausible analyst-controlled choices that preserve the defining structure of their respective protocols. We specified the perturbations before running the corresponding analyses. They were not selected in response to the displacement later observed.

The validity condition is strict: \textbf{the evaluated date--horizon pairs must be identical between the compared runs}. If the pairs differ, the contrast changes both the convention and the evaluation sample. It then measures sample replacement together with convention sensitivity rather than $N_{u,p}$. We verify pairing at the cell level before interpreting any displacement.

H2 is supported only when the between-protocol distribution exceeds the within-protocol references in paired comparisons. This result must hold for both the continuous measure and the decision-level measure.

\subsection{Protocol-sensitivity audit and computational cost}
\label{sec:audit}

The reusable audit has five steps:
\begin{enumerate}
\item Pre-specify the candidate set, evaluation units, evaluation coordinates, protocol configurations, and legitimate intraprotocol perturbations. Also pre-specify target identifiers, performance orientation, and tie handling.
\item Fit every candidate under every protocol configuration. Refit all models whenever the configuration changes.
\item Verify that each contrast intended to isolate a protocol or convention effect uses identical scored targets.
\item Compute $B_u$, each available $N_{u,p}$, and winner swap. Aggregate these quantities with uncertainty estimates that account for dependence among evaluation units.
\item Interpret claims of model superiority relative to the intraprotocol reference, not only relative to zero displacement.
\end{enumerate}

Constructing a reference requires at least one paired perturbation for each protocol. Additional defensible perturbations can produce a more informative reporting set. However, the number and range needed to characterise intraprotocol instability remain open questions.

Model fitting dominates the computational cost. With $K$ evaluated protocol/configuration combinations, the audit requires approximately $K|\mathcal{U}||\mathcal{M}|$ model-fit streams, including any rolling refits. Ranking and displacement calculations require only $O(K|\mathcal{U}|H M\log M)$ operations and are negligible relative to fitting.

\subsection{Panel inference}
\label{sec:stats}

We fixed the distinction between confirmatory and exploratory claims before analysis. We also fixed each threshold reported below. Supplementary Section~SM-02 documents these decisions.

\textbf{Confirmatory analysis.} At each station, we compute the Diebold--Mariano test \citep{diebold1995} with the finite-sample correction of \citet{harvey1997}, Kendall's $\tau$, and the Model Confidence Set \citep{hansen2011}.

At panel level, we report 95\% spatial block-bootstrap intervals for mean PSS and hard-reversal prevalence. The bootstrap uses $B=2000$ resamples. Countries define the blocks in Europe, and states define the blocks in the United States. This structure avoids treating neighbouring stations as independent information \citep{lahiri2003}.

We test equivalence of mean PSS with the two one-sided tests procedure (TOST; \citealp{schuirmann1987}). The equivalence margin is $\delta=0.10$, which is the rounded grid step from Equation~\ref{eq:grid-step} for $M=3$ and $H=7$.

Multiplicity control uses the Benjamini--Hochberg procedure \citep{benjamini1995}. The multiplicity family contains \textbf{one} station-level decision per site. In Europe, 8925 pairwise Diebold--Mariano tests underlie these decisions, but only 425 station-level decisions enter multiplicity control.

Power to detect a displacement of one grid step is 0.22. The analysis therefore has limited sensitivity to effects of that magnitude. Supplementary Section~SM-03 describes the exact decision construction and the alternative MCS-based severity analysis.

\textbf{Exploratory analysis.} We use Wilcoxon and sign tests to examine whether $\mathrm{PSS}>0$. We use Moran's $I$ with five nearest neighbours ($k=5$) to diagnose spatial dependence. Robustness checks vary the hard-reversal threshold and examine seasonality, sensitivity to imputation, and exclusion of 2020--2021. These analyses are descriptive and support no main claim.

\textbf{EPA panel.} We apply the same confirmatory battery separately to the US panel. Comparisons between the European and US panels are presented side by side rather than through pooled inference.

\subsection{Pre-commitment and instrument check}
\label{sec:frozen}
\label{sec:gate}

The confirmatory interpretation rests on a dated protocol freeze and a pre-panel instrument check. Supplementary Sections~SM-02 and SM-04 document these steps. No main result depends on the station used for the instrument check.

\subsection{Can series features predict sensitivity? (H5)}
\label{sec:index}

The diagnostic index fails its pre-specified viability gate: adjusted $R^2=0.105<0.15$ (Supplementary Section~SM-06). We therefore do not score the index on the EPA panel. Supplementary Sections~SM-08 and SM-07 report the secondary single-site and exceedance analyses. These analyses do not support H1--H5.

\section{Results}
\label{sec:results}

\subsection{How much rankings move between protocols (H1)}
\label{sec:h1}

The confirmatory panel contains 425 European background stations. Mean PSS is 0.801, with a 95\% block-bootstrap interval of 0.660 to 0.908 over 15 country blocks. The distribution is far from zero. Its median is 0.857, its interquartile range is 0.571 to 1.048, and its maximum is 1.619. The maximum is more than three-quarters of the distance to complete reversal. PSS is strictly positive at 420 of the 425 stations.

The equivalence margin provides a more concrete partition of the stations. Only 9 stations have $\mathrm{PSS}<\delta=0.10$ and are therefore practically insensitive to protocol choice. At the other extreme, 116 stations have $\mathrm{PSS}\ge 1$. At these stations, the two protocols do not even partially agree on the ranking. The remaining 300 stations fall between these two regions. At panel level, TOST does not establish equivalence to protocol insensitivity under $\delta=0.10$.

The decision-level result leads to the same conclusion. Hard reversal means that the winning model changes on a majority of the seven horizons. Its prevalence is 35.3\%, with an interval of 22.1\% to 47.6\%. Thus, at about one in three European background stations, changing the protocol does more than move the ranking. It changes the model that would be selected.

The significance-test results and the PSS results may initially seem inconsistent. Benjamini--Hochberg at $\alpha=0.10$, applied to one decision per station, rejects 83 of the 425 hypotheses of equal predictive accuracy. By contrast, PSS is positive at 420 stations. The two results are not contradictory because they answer different questions. The Diebold--Mariano test asks whether the difference in predictive accuracy is large enough, relative to the noise of one series, to be declared non-null. PSS asks whether the ranking changes when the protocol changes.

A ranking can reverse even when the differences in predictive accuracy are small. The observed data follow this pattern. Mean PSS is 0.624 at the 83 stations where the test rejects and 0.844 at the remaining 342 stations. The difference is supported by a Mann--Whitney test ($p=4.4\times 10^{-9}$). PSS is also negatively associated with the magnitude of the DM statistic (Spearman $\rho=-0.26$). Rankings are therefore more stable where the models differ more clearly, and instability is concentrated where their performance is closer.

This result is consistent with the limited power described in Section~\ref{sec:stats}. For series of this length, the probability of detecting a one-grid-step displacement is only 0.22. Much of the panel therefore lies in a difficult regime. The performance differences are sufficient to produce an ordering, but too small to survive a significance test. This is also the regime in which protocol choice has the greatest influence on the selected model. Consequently, for much of the panel, a superiority claim based on one protocol rests on differences that the formal test at the same station cannot distinguish from zero.

The exploratory spatial analysis also detects dependence. Moran's $I$ over the five nearest neighbours is 0.367, with simulated $p=0.005$. Nearby stations therefore tend to have similar PSS values. This result retrospectively supports the use of spatial blocks in the bootstrap. Treating all 425 stations as independent would produce intervals that are too narrow.

\textbf{Verdict H1: supported.} Rankings change between protocols across almost the entire panel. At about one-third of the stations, this movement also changes the selected model. The next section asks how much of this movement is specifically associated with the between-protocol contrast.

\begin{figure}[htbp]
\centering
\includegraphics[width=0.80\linewidth]{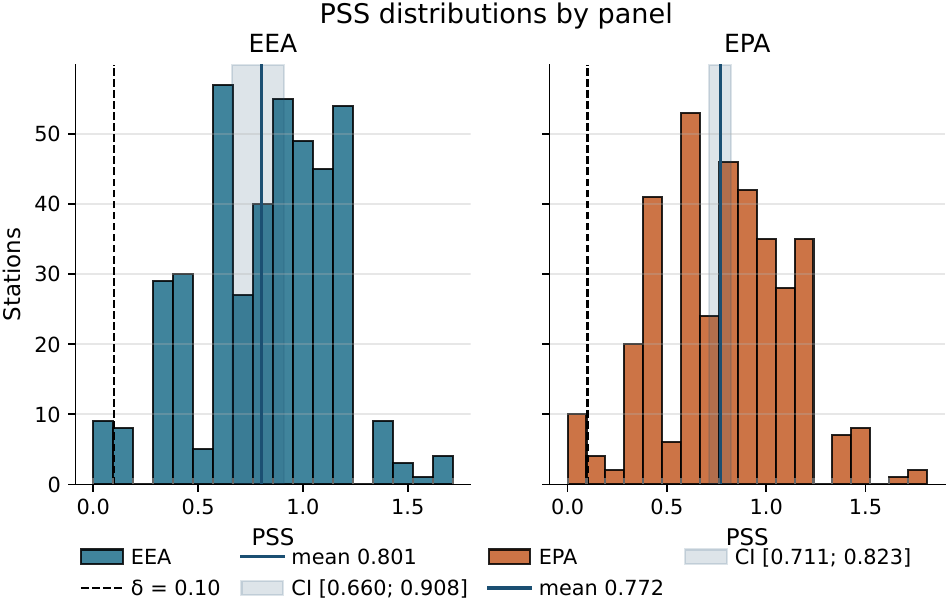}\\[0.6em]
\caption{PSS distributions on the European and US panels, with equivalence margin $\delta$ and the discrete measurement grid.}
\label{fig:pss}
\end{figure}

\begin{figure}[htbp]
\centering
\includegraphics[width=0.80\linewidth]{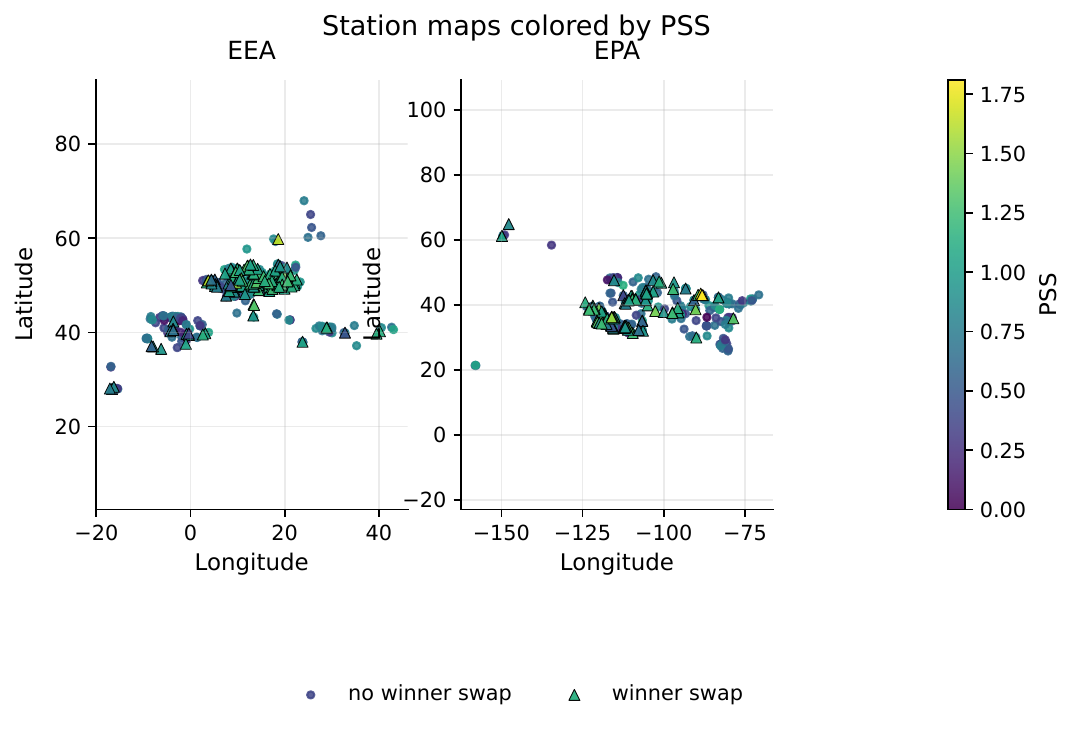}
\caption{Spatial maps of station-level PSS on the European and US panels.}
\label{fig:maps}
\end{figure}

Figure~\ref{fig:pss} shows the complete PSS distributions for the European and US panels. It also marks the equivalence margin $\delta$ and the discrete measurement grid. Figure~\ref{fig:maps} maps the station-level values and shows that sensitivity is not confined to one sub-region of either continent.

\subsection{Decomposition: how much movement belongs to the procedure (H2)}
\label{sec:h2}

A between-protocol PSS of 0.801 is difficult to interpret by itself. It requires a reference: how much would rankings move if the protocol remained fixed and only a conventional choice within it changed? Examples include the exact train boundary in a static split and the day of the month used for rolling-origin refitting. Several settings are defensible, but studies rarely report their effect on the ranking. If these within-protocol choices moved rankings as much as switching protocols, the result would indicate general procedure instability rather than a distinct protocol-choice effect.

We constructed two intraprotocol references. Both use full model refitting and the same evaluation window as the main analysis.

The first reference holds the static protocol fixed and shifts the train boundary by $\pm 1$ and $\pm 3$ months. The resulting displacement is small. Mean PSS is 0.072 in the European panel, with a 95\% interval of 0.064--0.083, and 0.065 in the US panel. The static split is therefore stable under the selected perturbations of its train boundary.

The second reference holds rolling-origin evaluation fixed and shifts the refit calendar by fifteen days. The compared runs preserve exactly the same evaluated date--horizon pairs. This pairing was verified cell by cell, so the measured displacement cannot be attributed to different evaluation samples. Under this perturbation, mean PSS is 0.230 in Europe, with an interval of 0.212--0.257, and 0.237 in the United States, with an interval of 0.197--0.277.

The between-protocol discrepancy is clearly larger than either reference. In Europe, the value of 0.801 exceeds the rolling-origin reference by an average paired difference of 0.571. Its interval is 0.425--0.674 and excludes zero. The between-protocol discrepancy is also above the placebo median at 96\% of stations and above its 95th percentile at 75\% of stations. Relative to the larger intraprotocol reference, switching protocols produces approximately three times as much ranking displacement.

The winner-swap measure gives an even sharper separation. Between protocols, the winner changes at 35.3\% of European stations. Under the static train-boundary perturbation, it changes at 0.8\% [0.4\%; 1.3\%]. Under the strictly paired rolling-origin calendar shift, it changes at 4.9\% [3.0\%; 8.1\%]. The US panel shows the same pattern: 31.5\% between protocols, compared with 1.2\% and 6.3\% under the two intraprotocol references. Table~\ref{tab:decomp} reports the corresponding intervals. The intervals are widely separated, and both PSS and winner swap support the same attribution.

The intraprotocol displacement is smaller, but it is not negligible. A PSS of 0.230 means that two legitimate runs of the same rolling-origin protocol produce noticeably different rankings when the refit calendar moves by two weeks. Studies rarely report this choice or treat it as a degree of freedom. Section~\ref{sec:implications} discusses the reporting consequence.

\textbf{Why target pairing is constitutive, not cosmetic.} An earlier rolling-origin placebo shifted the calendar without preserving the scored targets. It produced PSS 0.721, which was close enough to the between-protocol value to remove the attribution result. When the same conceptual perturbation preserves identical targets, PSS falls to 0.230. The nine-model arm independently reproduces this failure. A placebo with only 0.5\% coincident evaluation pairs yields 85.9\% winner swaps, compared with 22.8\% when the target pairs are identical. The unpaired designs therefore measure sample replacement together with convention sensitivity. They do not isolate intraprotocol convention noise. Supplementary Section~SM-05 reports the full distributions and controls.

\begin{table}[t]
\centering
\caption{Ranking-variance decomposition (H2) on both panels.}
\label{tab:decomp}
\begin{tabular}{@{}llccc@{}}
\toprule
Panel & Measure & Between protocols & Rolling convention & Static convention \\
\midrule
EEA & PSS & 0.801 [0.660; 0.908] & 0.230 [0.212; 0.257] & 0.072 [0.064; 0.083] \\
EEA & Winner swap & 35.3\% [22.1\%; 47.6\%] & 4.9\% [3.0\%; 8.1\%] & 0.8\% [0.4\%; 1.3\%] \\
EPA & PSS & 0.772 [0.711; 0.823] & 0.237 [0.197; 0.277] & 0.065 [0.044; 0.085] \\
EPA & Winner swap & 31.5\% [20.6\%; 38.9\%] & 6.3\% [3.4\%; 10.0\%] & 1.2\% [0.4\%; 2.1\%] \\
\bottomrule
\end{tabular}
\end{table}

\begin{figure}[!ht]
\centering
\includegraphics[width=0.80\linewidth]{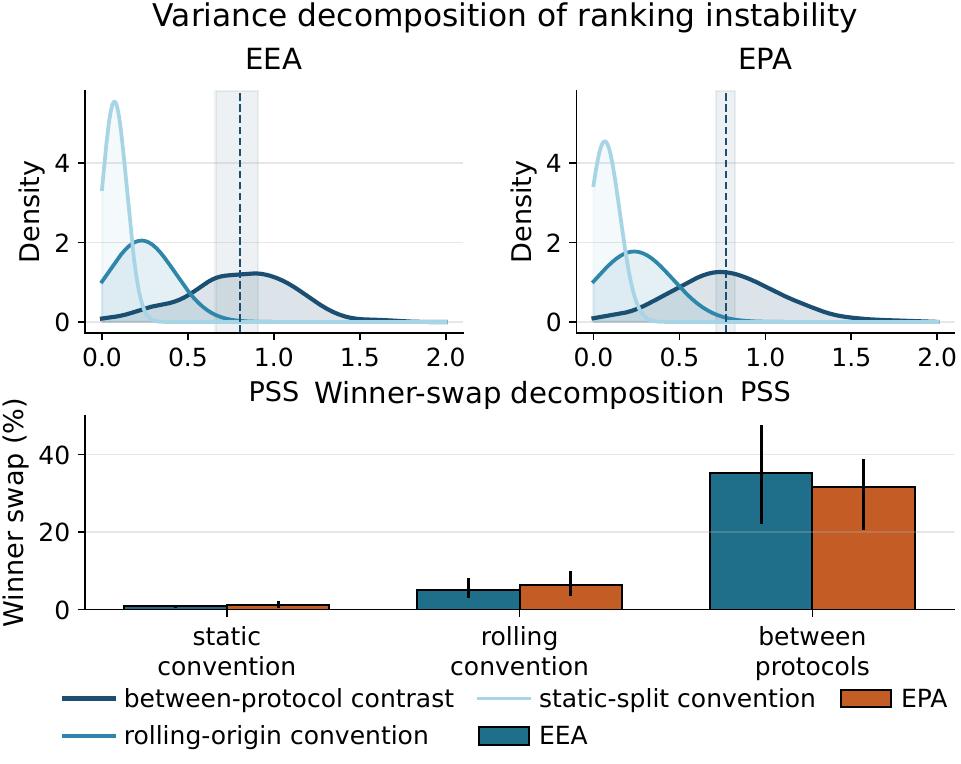}\\[0.6em]
\caption{Ranking-variance decomposition: between-protocol contrast versus intraprotocol nulls (PSS and winner-swap distributions).}
\label{fig:decomp}
\end{figure}

\textbf{Verdict H2: supported.} The observed between-protocol displacement is 11 times the static-split reference and 3.5 times the rolling-origin reference. The intervals are disjoint, and the PSS and winner-swap results agree. Table~\ref{tab:decomp} summarises the decomposition on both panels. Figure~\ref{fig:decomp} overlays the corresponding distributions. The next question is whether aggregating several legitimate runs of the same protocol can reduce this instability and what computational cost that would require.

\subsection{Where movement concentrates}
\label{sec:where}

Three exploratory analyses show where the measured discrepancy is concentrated.

\textbf{Discriminatory power of the protocols.} The confidence sets differ substantially in width. Under rolling-origin evaluation, the Model Confidence Set contains all three models at 65.3\% of station--horizon pairs, and its mean size is 2.50. Under the static split, it contains all three models at only 9.1\% of pairs, and its mean size is 1.79. Thus, the protocol that more closely represents repeated operational updating is also the protocol that least clearly separates the models.

\textbf{Where rankings move.} The most common outcome is apparent stability: XGBoost leads under both protocols at 55.5\% of European stations. Under rolling-origin evaluation, persistence becomes the leader at 8.0\% of European stations and 11.8\% of US monitors. The specific transition reported in the illustrative case---from XGBoost to persistence---occurs at only 0.9\% of the European panel (Supplementary Section~SM-08). The general instability pattern therefore extends across the panel, but this particular direction of change does not.

\textbf{How discrepancy varies with horizon.} The results do not support the intuition that longer horizons should be more unstable. PSS is highest at the one-day horizon, where it reaches 0.910, and declines to approximately 0.591 at the longest horizon. At $h=1$, the winner-disagreement rate is 47.3\%. In this case study, protocol choice matters most at the horizon with the greatest operational relevance.

\subsection{Decision space: more candidates, less stability (H3)}
\label{sec:nine}

The preceding analyses compare three models, whereas applied studies often compare larger candidate sets. One possible expectation is that a larger set would stabilise selection because the best model would have more opportunity to stand out. We test this possibility by repeating both panels with nine models. The added candidates are seasonal naive, ETS, Theta, a regularised autoregressive method, a second tree method, and automatic ARIMA order selection.

We re-derived the equivalence margin using the same rule as in the three-model arm, $\delta(M)=4/[M(M-1)H]$. The margin therefore changes from 0.095 to 0.0079. Under the new margin, TOST establishes equivalence on neither panel. We also rebuilt the intraprotocol references using the same nine-model set and strict pairing of the evaluated date--horizon pairs (Table~\ref{tab:nine}).

\begin{table}[t]
\centering
\caption{Nine-model arm: between-protocol contrast vs intraprotocol placebos.}
\label{tab:nine}
\begin{tabular}{@{}llccc@{}}
\toprule
Panel & Measure & Between protocols & Static placebo & Rolling placebo \\
\midrule
EEA & PSS & 0.540 [0.499; 0.572] & 0.157 [0.148; 0.165] & 0.251 [0.233; 0.276] \\
EEA & Winner swap & 60.2\% [56.0; 64.3] & 27.2\% [18.8; 35.0] & 22.8\% [17.5; 30.3] \\
EEA & MCS exclusion & 1.4\% [0.4\%; 3.4\%] & 1.2\% [0.5\%; 2.5\%] & 4.7\% [3.0\%; 7.5\%] \\
EPA & PSS & 0.550 [0.505; 0.583] & 0.169 [0.153; 0.181] & 0.273 [0.251; 0.294] \\
EPA & Winner swap & 67.4\% [61.3; 73.3] & 16.2\% [12.3; 21.6] & 27.4\% [21.6; 32.7] \\
EPA & MCS exclusion & 1.4\% [0.4\%; 2.7\%] & 0.4\% [0.1\%; 0.9\%] & 4.4\% [2.1\%; 7.9\%] \\
\bottomrule
\end{tabular}
\end{table}

\noindent $\delta(M=9)=0.0079$; TOST not equivalent on both panels; pairing fraction $=1.0$.

\textbf{Both measures continue to support attribution.} On both panels, the intervals for the between-protocol results are disjoint from the intervals for both intraprotocol placebos. Relative to the larger intraprotocol reference, the ratio is 2.2 in Europe and 2.0 in the United States. These ratios are smaller than the value of 3.5 in the three-model arm, but the separation remains clear.

\textbf{The two measures respond differently to the larger candidate set.} In Europe, between-protocol winner swaps increase from 35.3\% to 60.2\%, whereas mean PSS decreases from 0.801 to 0.540. These results are compatible. With nine candidates, many pairwise relations remain unchanged: persistence stays near the bottom and the more flexible methods stay near the top. Kendall's $\tau$ weights all candidate pairs equally, so these stable pairwise relations reduce the average displacement. The competition is concentrated near the top of the ranking, where the final selection is made. The distinction between full-ranking displacement and Top-1 choice change, introduced in Section~\ref{sec:protocols}, is therefore especially important in the expanded arm.

The intraprotocol reference also becomes less stable. Under the rolling-origin perturbation, winner swaps increase from 4.9\% with three models to 22.8\% with nine. Both the between-protocol displacement and the within-protocol reference therefore increase at the decision level. Their separation narrows but remains present. Within the candidate-set expansion examined here, model selection becomes more sensitive both to protocol choice and to the selected intraprotocol conventions.

\textbf{The MCS-exclusion criterion still lacks useful power.} With three models, this criterion is nearly inert, with a prevalence of 0.24\%. Adding candidates was expected to make the Model Confidence Set more discriminatory. The result does not support that expectation. With nine models, prevalence increases only to 1.4\% [0.4\%; 3.4\%]. Under the rolling-origin placebo, it reaches 4.7\% [3.0\%; 7.5\%], which is higher than the between-protocol rate. In both candidate sets, the criterion operates on only a few dozen stations and does not separate the effect. We interpret this as a limitation of the instrument, not as evidence against the phenomenon. For series of this length, the confidence sets are too wide to support an exclusion-based severity criterion. PSS and winner swap continue to characterise the effect concordantly.

\begin{figure}[htbp]
\centering
\includegraphics[width=0.80\linewidth]{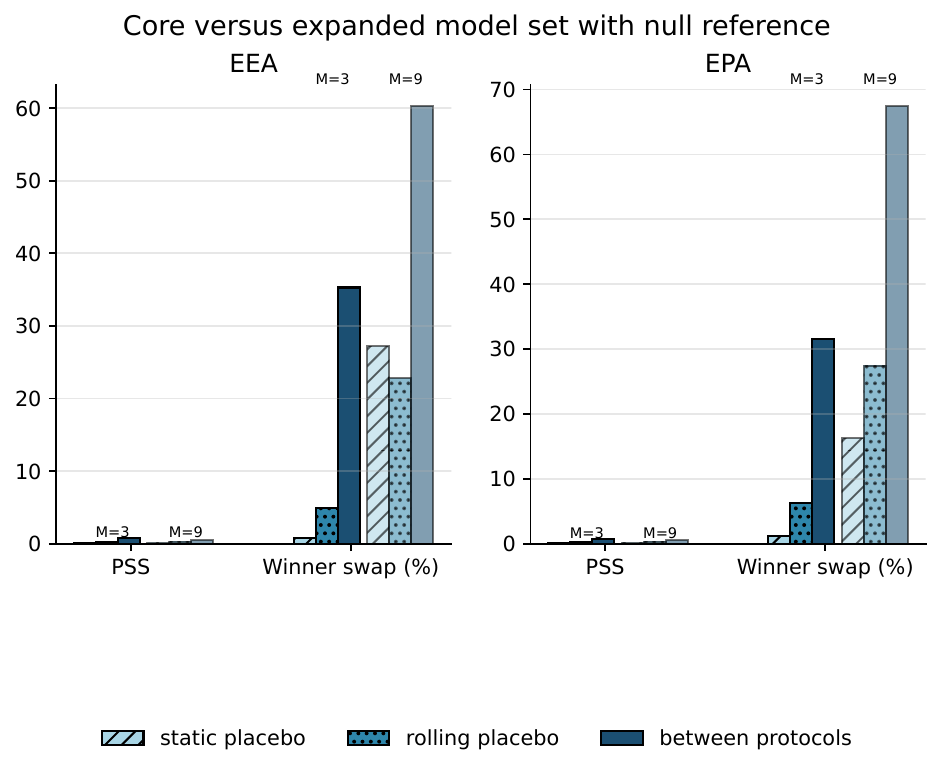}
\caption{Three-model versus nine-model decision spaces: between-protocol contrast and intraprotocol nulls under PSS and winner swap.}
\label{fig:corexp}
\end{figure}

Table~\ref{tab:nine} reports the complete expanded-arm battery. Figure~\ref{fig:corexp} compares the three- and nine-model sets under both measures and their intraprotocol references.

\textbf{Verdict H3: supported within the candidate-set expansion examined here.} With nine models, the between-protocol contrast remains separated from the selected intraprotocol references on both continents and under both measures supported by the design. Selection stability is lower in this expanded arm. Winner swaps rise from 35.3\% to 60.2\% in Europe and from 31.5\% to 67.4\% in the United States. Comparative studies often evaluate increasingly large candidate sets, so this result may matter beyond air-quality forecasting. Its validation outside this case study remains open.

\subsection{Run aggregation and selection stability}
\label{sec:agg}

Section~\ref{sec:h2} asks whether aggregating several intraprotocol runs can reduce the between-protocol winner-swap rate. We expand the intraprotocol grids on both panels. For the static protocol, the train boundary shifts by $\pm1,\ldots,\pm6$ months, giving 13 runs including the reference. For rolling-origin evaluation, the refit calendar shifts by $\pm3,\ldots,\pm21$ days in increments of three days, giving 15 runs. We retain only perturbations that preserve exact scored $({\rm date},h)$ pairs relative to the reference. This rule discards $\AggStabNDisc$ cells.

\begin{figure}[htbp]
\centering
\includegraphics[width=1.0\linewidth]{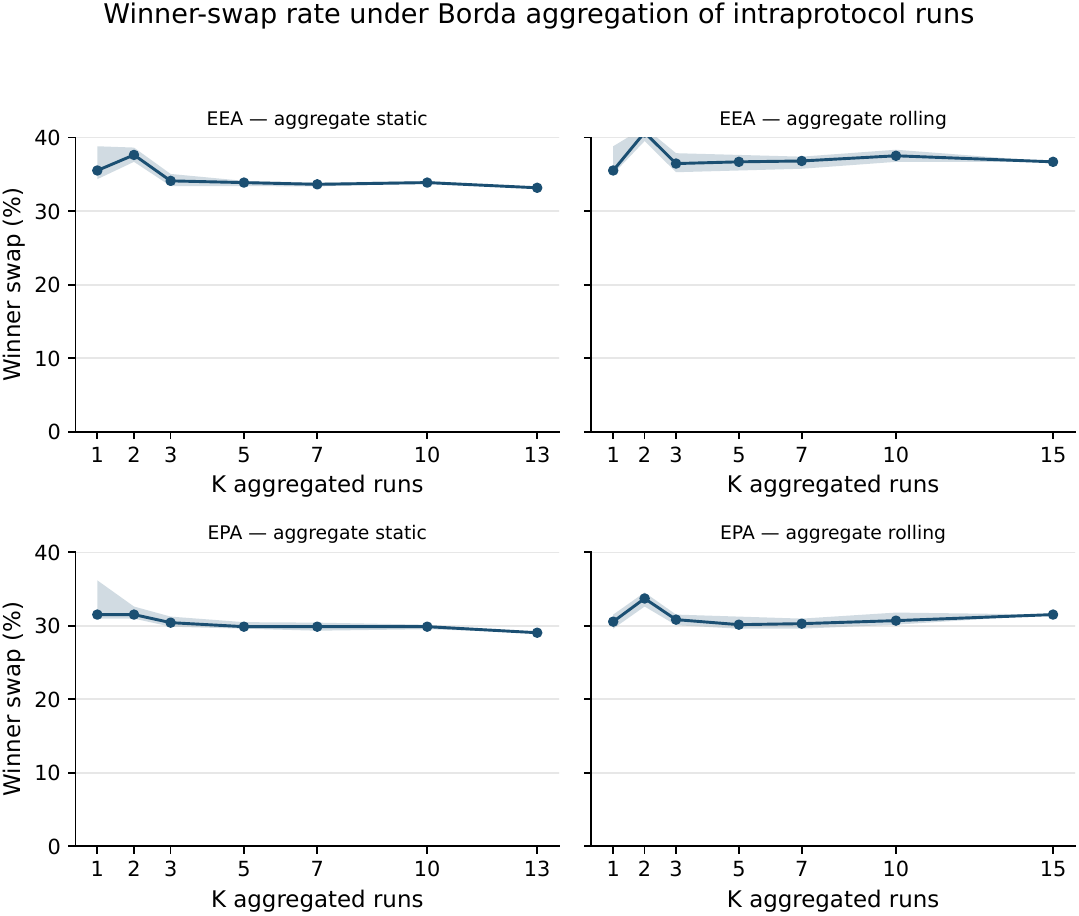}
\caption{Winner-swap rate versus number of Borda-aggregated intraprotocol runs ($K$). Rows: EEA and EPA panels. Columns: Curve~A (aggregate static, compare to rolling reference) and Curve~B (aggregate rolling, compare to static reference). Solid line: median over 100 random subsets; band: interquartile range.}
\label{fig:agg}
\end{figure}

For each value of $K$, we draw 100 random subsets of runs. We aggregate their rankings using the Borda mean rank and break ties alphabetically. We then calculate hard-reversal prevalence relative to the opposite protocol's reference run. Figure~\ref{fig:agg} and Table~\ref{tab:agg} report two curves. Curve~A aggregates runs from the static arm and compares them with the rolling reference. Curve~B aggregates rolling-origin runs and compares them with the static reference. Supplementary Section~SM-10 reports a joint aggregation of both arms.

\begin{table}[t]
\centering
\caption{Winner-swap rate (\%) under Borda aggregation: median [IQR] over 100 subsets.}
\label{tab:agg}
\begin{tabular}{@{}llccc@{}}
\toprule
Panel & Curve & $K$ & Median [\%] & IQR [\%] \\
\midrule
EEA & Static agg. & 1 & 35.5 & [34.4; 38.8] \\
EEA & Static agg. & 3 & 34.1 & [33.4; 35.1] \\
EEA & Static agg. & 5 & 33.9 & [33.4; 34.1] \\
EEA & Static agg. & 7 & 33.6 & [33.4; 33.9] \\
EEA & Static agg. & 10 & 33.9 & [33.8; 34.1] \\
EEA & Static agg. & 13 & 33.2 & [33.2; 33.2] \\
EEA & Rolling agg. & 1 & 35.5 & [35.1; 38.8] \\
EEA & Rolling agg. & 3 & 36.5 & [35.3; 37.9] \\
EEA & Rolling agg. & 5 & 36.7 & [35.5; 37.6] \\
EEA & Rolling agg. & 7 & 36.8 & [35.8; 37.4] \\
EEA & Rolling agg. & 10 & 37.5 & [36.7; 38.4] \\
EEA & Rolling agg. & 15 & 36.7 & [36.7; 36.7] \\
EPA & Static agg. & 1 & 31.5 & [31.0; 36.2] \\
EPA & Static agg. & 3 & 30.4 & [29.9; 31.2] \\
EPA & Static agg. & 5 & 29.9 & [29.6; 30.5] \\
EPA & Static agg. & 7 & 29.9 & [29.3; 30.4] \\
EPA & Static agg. & 10 & 29.9 & [29.5; 30.1] \\
EPA & Static agg. & 13 & 29.0 & [29.0; 29.0] \\
EPA & Rolling agg. & 1 & 30.5 & [29.6; 31.5] \\
EPA & Rolling agg. & 3 & 30.8 & [30.1; 31.5] \\
EPA & Rolling agg. & 5 & 30.1 & [29.6; 31.2] \\
EPA & Rolling agg. & 7 & 30.3 & [29.6; 31.0] \\
EPA & Rolling agg. & 10 & 30.7 & [30.1; 31.8] \\
EPA & Rolling agg. & 15 & 31.5 & [31.5; 31.5] \\
\bottomrule
\end{tabular}
\end{table}

On the European panel, Curve~A changes only from $\AggStabFromA\%$ at $K=1$ to a plateau of $\AggStabToA\%$. Curve~B remains near $\AggStabFromB$--$\AggStabToB\%$. The single-run hard-reversal baseline is $\AggStabWSBase\%$. Both curves are essentially stable by $K=\AggStabKRec$. The EPA panel shows the same qualitative pattern, with plateaus of $\AggStabEPAPlateauA\%$ and $\AggStabEPAPlateauB\%$.

Aggregation therefore does not remove the between-protocol discrepancy. The residual gap, defined as the mean of the two European plateaus, remains $\AggStabScaleX$ times larger than the rolling-origin intraprotocol reference in Table~\ref{tab:decomp}. Protocol choice and the selected convention noise therefore continue to operate at different scales after aggregation. This calibration informs the ranking-variance report proposed in Section~\ref{sec:implications}. A small number of legitimate intraprotocol reruns is sufficient to measure the residual gap, but not to eliminate it.

\subsection{Confirmation on reserved material and non-background strata (H4, first part)}
\label{sec:holdout}

The preceding European analyses use the background stratum, and some criteria were refined during the study. We therefore evaluated the main claim on material that these decisions had not examined. Before selecting this material, we froze the complete protocol specification: models, horizons, PSS and winner-swap definitions, statistical plan, and block definition.

After the freeze, we selected eligible Heisig stations by a purely structural rule. The confirmation material contains 271 remaining background stations and 520 traffic or industrial stations. None of their metrics had been calculated at the time of selection. We ran the frozen protocol once and made no subsequent adjustments.

\begin{table}[t]
\centering
\caption{Frozen-protocol confirmation sets.}
\label{tab:holdout}
\begin{tabular}{@{}lcccc@{}}
\toprule
Stratum & $N$ & Blocks & Mean PSS [95\% CI] & Winner swap [95\% CI] \\
\midrule
Background confirmatory & 425 & 15 & 0.801 [0.660; 0.908] & 35.3\% [22.1\%; 47.6\%] \\
Background holdout & 271 & 4 & 0.676 [0.606; 0.833] & 20.3\% [17.9\%; 37.5\%] \\
Traffic / industrial & 520 & 19 & 0.701 [0.617; 0.847] & 23.1\% [16.5\%; 32.8\%] \\
\bottomrule
\end{tabular}
\end{table}

Table~\ref{tab:holdout} compares the confirmatory panel with the two reserved sets. Mean PSS for the reserved background stations lies inside the interval of the confirmatory panel, and the intervals for all three sets overlap widely. Winner-swap prevalence is lower in the reserved sets, although their intervals still touch the confirmatory interval.

The traffic and industrial stations address an important limitation of the original design. Their local emission dynamics could alter both predictability and sensitivity to the evaluation protocol. Nevertheless, their mean PSS is 0.701, and its interval overlaps widely with the background interval. Within this case study, the between-protocol discrepancy is therefore not confined to the background site type.

Two qualifications remain. First, no metric had been computed on the 271 reserved background stations before the freeze, apart from one declared exception documented in Supplementary Sections~SM-02 and SM-04. Second, these 271 stations are distributed across only four countries. The country-block interval therefore relies on four blocks. The point estimate and verdict remain stable under finer spatial definitions using 300\,km and 500\,km bands and under an analysis that treats stations as independent, which gives an interval of [0.643; 0.713].

\textbf{Verdict H4, first part: supported.} The main pattern remains after a single frozen-protocol run on previously unexamined material. It also appears in traffic and industrial strata that were excluded from the original background design.

\subsection{Cross-continental transfer (H4, second part)}
\label{sec:transfer}

We applied the identical protocol to 365 EPA monitors. Mean PSS is 0.772, with a block-bootstrap interval of 0.711 to 0.823 over 38 state blocks. Hard-reversal prevalence is 31.5\%, with an interval of 20.6\% to 38.9\%. PSS is positive at 362 of 365 monitors, and TOST again does not establish equivalence under $\delta=0.10$. The US distribution also resembles the European distribution in shape. Its median is 0.762, its interquartile range is 0.571 to 0.952, and its maximum is 1.810.

The difference in mean PSS between the panels is $-0.029$. This is smaller than the grid step of the measure, $2/21\approx 0.095$, and therefore smaller than the smallest change that the instrument can distinguish in the three-model arm. Independent block bootstraps within the two panels place this descriptive difference in $[-0.155; 0.124]$. In 74.8\% of resamples, its magnitude is below the grid step.

We do not perform a formal test of the difference between panels. Their sampling designs, block structures, and source networks differ. A formal test could imply a level of statistical equivalence that the study was not designed to establish. The comparison is therefore descriptive.

The EPA panel also shows short-range spatial dependence. Moran's $I$ is 0.223 under five nearest neighbours, with simulated $p=0.005$. It decreases to 0.105 under 300\,km bands and to 0.062 under 500\,km bands. This attenuation is compatible with a short-range pattern similar to that observed in Europe.

\textbf{Verdict H4, second part: supported.} The instability pattern appears in an independent network on another continent under the same protocol. The panels remain separate, and the diagnostic index is not transferred.

\subsection{Confirmatory battery summary}
\label{sec:summary}

Table~\ref{tab:battery} presents the two confirmatory panels side by side. They remain separate in all analyses. The intervals have different widths because the European analysis uses 15 country blocks and the US analysis uses 38 state blocks. The difference does not represent different precision of the station-level measures.

\begin{table}[t]
\centering
\caption{Side-by-side confirmatory battery.}
\label{tab:battery}
\begin{tabular}{@{}lccccccc@{}}
\toprule
Panel & $N$ & Blocks & Mean PSS & Winner swap & PSS$>$0 & TOST ($\delta=0.10$) & BH-FDR \\
\midrule
EEA & 425 & 15 & 0.801 [0.660; 0.908] & 35.3\% [22.1; 47.6] & 420/425 & not equiv. & 83 \\
EPA & 365 & 38 & 0.772 [0.711; 0.823] & 31.5\% [20.6; 38.9] & 362/365 & not equiv. & -- \\
\bottomrule
\end{tabular}
\end{table}

\subsection{Complementary single-site module}
\label{sec:singlesite}

Elche is in the most stable PSS decile. Its PSS is 0.286, and it has no winner swap. Even at this comparatively stable station, the protocol-to-noise ratio remains approximately three to one. Supplementary Section~SM-08 reports the complete analysis.

\subsection{Diagnostic index (H5)}
\label{sec:h5}

The diagnostic index does not pass the pre-specified viability gate. Its adjusted $R^2$ is 0.105, below the threshold of 0.15 (Supplementary Section~SM-06).

\textbf{Verdict H5: not supported.}

Supplementary Section~SM-07 reports the exceedance results.

\section{Discussion}
\label{sec:discussion}

\subsection{Reference-based attribution and its interpretation}
\label{sec:decomp-disc}

The central result is a reference-based attribution. On both continents and under both measures, the displacement between protocols is substantially larger than the displacement produced by the selected intraprotocol perturbations (Section~\ref{sec:h2}). This comparison gives the observed between-protocol difference an empirical reference.

Two alternative explanations were examined within the sequential case study. The protocols use identical scored targets, so unequal test sets cannot explain the discrepancy. A contiguous-training variant also indicates that training-data lag does not account for most of the observed difference. These checks narrow the interpretation, but they do not turn the analysis into an exhaustive decomposition of every possible protocol convention.

The intraprotocol displacement is smaller, but it is not negligible. Under this study's equivalence criterion, two legitimate runs of the same protocol are not equivalent. In particular, shifting the rolling-origin refit calendar by fifteen days produces mean PSS of 0.230. The day on which refitting occurs is rarely reported, yet it can affect the resulting ranking.

The aggregation analysis in Section~\ref{sec:agg} examines whether the main discrepancy is mainly a consequence of choosing one intraprotocol run. Across the perturbation grids studied here, winner-swap rates stabilise after only a few aggregated runs and remain close to the single-run between-protocol rate. Aggregating the examined runs therefore does not remove the discrepancy. This result supports the interpretation that the between-protocol contrast operates at a larger scale than the selected convention changes. It does not establish a universal distinction between structural and stochastic variation.

Two additional analyses define how far this interpretation extends within the case study. First, the pattern reappears in 791 stations that were not used in the original inference, including traffic and industrial stations. This shows that the result is not confined to the original background panel. Second, the separation from the intraprotocol references remains after expanding the candidate set from three to nine models. In that expansion, winner swaps increase from 35.3\% to 60.2\%. The added candidates create a denser contest near the top of the ranking, where model selection occurs. Whether the same pattern holds for substantially larger candidate sets remains untested.

The specific transition reported in the single-site case is rare across the panel ($<1\%$). What generalises within this study is the presence of ranking displacement, not the outcome of that individual station (Supplementary Section~SM-08).

\subsection{Robustness checks and alternative explanations}
\label{sec:refutations}

We examined the main result against several plausible alternative explanations. Each explanation was linked to an executable analysis and an interpretation rule specified before the corresponding test. Table~\ref{tab:refute} summarises these checks.

\begin{table}[t]
\centering
\caption{Attempted alternative explanations and outcomes.}
\label{tab:refute}
\small
\begin{tabular}{@{}p{3.2cm}p{5.5cm}p{4.5cm}@{}}
\toprule
Alternative & How tested & Outcome \\
\midrule
Single-station artefact & Panels of 425 European and 365 US monitors, independent networks, identical protocol, no pooling & PSS positive at 420/425 and 362/365 (Sections~\ref{sec:h1}, \ref{sec:singlesite}; Supplementary Section~SM-08) \\
Evaluation-procedure noise & Full-refit intraprotocol references with strict target pairing & PSS 0.801 between protocols versus 0.072 and 0.230 within protocols; winner swaps 35.3\% versus 0.8\% and 4.9\% (Section~\ref{sec:h2}) \\
Unequal test sets & Count of date--horizon pairs per station under each protocol & Counts are identical on both panels (Section~\ref{sec:protocols}) \\
Static-split training lag & European re-run with training contiguous to evaluation & PSS 0.773 versus 0.801; mean absolute difference 0.073, below the grid step (Section~\ref{sec:limitations}) \\
Background-stratum restriction & Frozen protocol on 520 traffic and industrial stations & Mean PSS 0.701, with an interval that overlaps the background interval (Section~\ref{sec:holdout}) \\
Adaptation to examined data & Dated freeze before selecting 271 previously unexamined stations; one run & Mean PSS 0.676, inside the confirmatory interval (Section~\ref{sec:holdout}) \\
Small model-set artefact & Full repetition with nine models and a re-derived margin & Discrepancy-to-reference ratio approximately 2 under both measures and on both panels (Section~\ref{sec:nine}) \\
Underestimated spatial dependence & Intervals recomputed under four block definitions & Estimates remain in the same order of magnitude (Section~\ref{sec:limitations}) \\
\bottomrule
\end{tabular}
\end{table}

The target-pairing diagnostic in Section~\ref{sec:h2} changed the interpretation during study development. It therefore belongs to the scientific result rather than only to the implementation history. When the intended estimand is sensitivity to a convention, the compared runs must preserve the scored examples. Otherwise, the measured displacement combines convention variation with replacement of the evaluation sample. Supplementary Section~SM-05 reports the extended diagnostics.

\subsection{What the index does not show, and why it matters}
\label{sec:index-disc}

The diagnostic index did not pass its pre-specified viability gate (Section~\ref{sec:h5}). Under the tested specification, the observable series features explain too little variation to identify stations with greater ranking instability.

This negative result has a practical implication. The tested features cannot be used as a reliable shortcut for deciding where a protocol-sensitivity audit is needed. Within the present study, sensitivity must therefore be assessed directly by comparing protocols and their selected perturbations. The result does not show that instability is inherently unpredictable. Other predictors, including features of the models' error regimes, remain possible directions for future work.

\subsection{Limitations}
\label{sec:limitations}

The scope of the claims differs across five axes:
\begin{enumerate}
\item \textbf{Geography --- validated within the case study.} The analysis covers independent networks on two continents under the same protocol. The panels are never pooled.
\item \textbf{Site type --- extended within the case study.} The pattern also appears at traffic and industrial stations (Section~\ref{sec:holdout}). Their interval overlaps the background interval, but the study does not establish formal equivalence among site types.
\item \textbf{Decision space --- examined up to nine models.} Between-protocol displacement exceeds the selected intraprotocol references with both three and nine candidates, under both measures and on both panels. The ratio narrows from 3.5 to approximately 2 in the nine-model arm (Section~\ref{sec:nine}). Behaviour in substantially larger candidate sets is unknown.
\item \textbf{Pollutant and frequency --- not tested.} The empirical claims are limited to \textbf{daily PM10}. Series with different autocorrelation or seasonality may behave differently.
\item \textbf{Domains beyond sequential prediction --- not validated.} The proposed methodology is domain-agnostic, but the empirical validation covers only multi-step daily-PM10 prediction under static-split and rolling-origin protocols. Applications to i.i.d.\ or covariate-shift classification, tabular AutoML, computer vision, hyperparameter optimisation, medical prediction, and foundation-model evaluation remain future work.
\end{enumerate}

Several additional limitations concern the design and the measures. Pipeline calibration was verified conceptually rather than numerically, and no main claim depends on that check. The observation window ends in 2023 and includes the 2020--2021 shock. Excluding that period is planned as a robustness analysis, but the current conclusions do not depend on such an exclusion.

PSS also has limited resolution in the three-model arm. With three models and seven horizons, it lies on a grid with step $\approx 0.095$. The estimated power to detect a one-step displacement is 0.22, so the design has limited sensitivity to effects of that size. In addition, the aggregation analysis in Section~\ref{sec:agg} uses only Borda mean rank. Alternative social-choice aggregation rules were not compared.

\textbf{Winner swap is a counting rule rather than a per-station significance test.} It records whether the top-ranked model differs on at least four of seven horizons. It does not require that this difference be statistically significant at the individual station. We also tested an inference-anchored severity criterion based on MCS exclusion, but it was practically inert in both model-set sizes (Supplementary Section~SM-03).

The panel-level interpretation therefore relies on the intraprotocol reference distributions. The same winner-swap rule is applied to the same series and models under perturbations that preserve the protocol. The resulting reference intervals are well separated from the between-protocol intervals in both model sets and on both continents. This empirical reference answers a panel-level question: whether the observed frequency is larger than the frequency produced by the selected convention changes. It does not provide a per-station error theory or determine whether any individual winner swap is noise.

The design rules out unequal evaluation samples as an explanation for the main contrast. Both protocols score the same date--horizon pairs, at the same monthly origins, over the same two-year window. Their counts are identical on both panels.

Training-data recency is a more serious potential explanation because the original protocols differ in both evaluation structure and the age of the training data (Section~\ref{sec:protocols}). We therefore repeated the European analysis with a static-split variant whose training period ends immediately before evaluation. This removes the training gap while preserving the remaining design. Mean PSS changes from 0.801 to 0.773, with an interval of 0.638 to 0.883. The mean paired difference is 0.028, the mean absolute difference is 0.073, and the ratio to the larger internal reference remains 3.4 to 1. In this variant, training recency does not account for most of the observed between-protocol displacement. The analysis does not exclude every possible interaction between training history and protocol design.

The single-site module has restricted scope and does not alter the panel attribution in Section~\ref{sec:h2} (Supplementary Section~SM-08).

Residual spatial dependence remains on both panels. We therefore recomputed intervals under alternative block definitions: 300\,km bands, 500\,km bands, and k-nearest-neighbour neighbourhoods. The estimates remain in the same order of magnitude. For example, the European mean-PSS intervals are 0.712--0.886 and 0.667--0.896 under two alternatives, compared with 0.660--0.908 under country blocks. The conclusion is therefore not dependent on one block definition. Finally, the panels are not pooled, and their comparison remains descriptive. The study was not designed to establish statistical equivalence between continents.

\subsection{Implications}
\label{sec:implications}

The implications have two levels. The first concerns the choice between protocols. In this sequential case study, a superiority claim based on only one protocol is conditional on that protocol. An equally legitimate alternative protocol often produces a different ranking. This does not show that one protocol is correct and the other is wrong. It shows that the selected model can depend on a methodological condition that comparative studies do not always report explicitly.

The second level concerns conventions within one protocol. Shifting the rolling-origin refit calendar by fifteen days produces mean PSS of 0.230. Choosing a protocol with desirable bias properties does not remove this source of ranking variation. In an isolated evaluation run, such variation can affect which model is selected.

These results motivate a concrete reporting practice: a \textbf{ranking-variance report}. The report has three steps. First, repeat the evaluation under a small set of legitimate intraprotocol perturbations. In this case study, examples are shifting the refit calendar by one or two weeks and moving the training boundary by one or three months. Second, report the displacement of the ranking and how often the winning model changes across these runs. Third, interpret any difference between evaluation procedures relative to this intraprotocol reference.

The aggregation analysis in Section~\ref{sec:agg} provides an initial calibration for this recommendation. Aggregating $K=\AggStabKRec$ runs leaves the European between-protocol winner-swap rate near $\AggStabToA$--$\AggStabToB\%$. This is close to the single-run baseline of $\AggStabWSBase\%$ and remains $\AggStabScaleX$ times the rolling intraprotocol reference. In this study, a few perturbations are therefore sufficient to reveal that the between-protocol gap remains larger than the selected convention noise. The required computation is proportional to the number and cost of the additional refits; it may not be small for every model family or dataset.

The nine-model arm in Section~\ref{sec:nine} makes this reporting step more relevant, not less. In the candidate-set expansion examined here, between-protocol winner swaps increase from 35.3\% to 60.2\%, and intraprotocol winner swaps also increase. This result does not establish a monotonic relationship between candidate-set size and instability. It shows that adding candidates did not stabilise selection in this particular expansion.

The same methodological question can be posed in other areas of empirical machine learning: how much of a ranking is associated with protocol choice, and how much is observed under conventions within the protocol? Possible settings include cross-validation design, AutoML comparisons, hyperparameter studies, and foundation-model evaluation harnesses. These are proposed applications of the framework, not empirical results of the present study.

The analysis does not identify a universally superior predictive model, nor is that its purpose. Its contribution concerns the procedure through which comparative claims and model-selection decisions are produced.

\subsection{Future work}
\label{sec:future}

Future work should first test the axes on which the present claims remain unverified.

\textbf{A severity measure with useful power.} The MCS-exclusion criterion is not useful in this design. It is nearly inert with three models and reaches only 1.4\% with nine models, below the rate produced by its rolling-origin reference. A stronger criterion should remain anchored in statistical inference while retaining power for series of this length. Possible approaches include aggregation across horizons and confidence sets constructed over multiple runs.

\textbf{Variance reduction through aggregation.} Section~\ref{sec:agg} provides an initial analysis using Borda aggregation over the intraprotocol grids. Future work should compare aggregation rules and determine how many perturbations are needed to reduce residual instability at an acceptable computational cost. The nine-model arm should also be included in that analysis.

\textbf{Additional axes within air-quality prediction.} The same networks contain NO$_2$ and PM2.5 under the common eligibility criteria. These pollutants would allow a test of whether protocol sensitivity changes with the predictability of the series. The reserved background material also spans only four countries. Repeating the freeze--select--run design on a more geographically dispersed eligible universe (Section~\ref{sec:eea}) would provide a broader confirmation.

\textbf{External domains.} The reference-based decomposition can be evaluated wherever predictive models are compared under alternative legitimate protocols. Candidate settings include classification under different resampling schemes, tabular AutoML, computer-vision benchmarks, hyperparameter optimisation, medical prediction, and foundation-model evaluation. These are potential applications, not validated claims of this study.

Two further directions arise from the secondary analyses. First, the failure of the diagnostic index suggests that the marginal series features tested here are insufficient. Features of the models' errors or of their relative performance may be more informative. Second, the single-site module suggests a possible relationship between data completeness and ranking stability. The original transition appears under a restrictive completeness rule but disappears under a more permissive rule. This makes gap treatment a candidate factor for future investigation, not an established mechanism.

\section{Conclusion}
\label{sec:conclusion}

This study asked how much a model-selection conclusion reflects the models being compared and how much it depends on the evaluation protocol.

We examined this question in a sequential-prediction case study of daily PM10. The analysis includes 425 European background stations and 365 US EPA monitors. Changing from static-split to rolling-origin evaluation frequently changes the model ranking. The between-protocol displacement is also substantially larger than the displacement produced by the selected conventions within either protocol. Depending on the intraprotocol reference, the difference is approximately three to seven times larger.

Two checks narrow the interpretation of this result. The protocols score identical date--horizon pairs, so unequal test sets do not explain the difference. A contiguous-training variant also indicates that the training-data gap in the original static split does not account for most of the observed displacement.

The pattern remains under a frozen-protocol run on 271 previously unexamined stations and appears in 520 traffic and industrial stations. It also remains after expanding the candidate set from three to nine models. In this expansion, the proportion of stations with a between-protocol winner swap increases from 35.3\% to 60.2\%. This result shows that adding candidates did not stabilise selection in the model set examined here. It does not establish that instability must increase with candidate-set size in general.

These findings change how superiority claims should be interpreted in this setting. The issue is not that one of the two protocols must be rejected. Both are legitimate evaluation choices. The issue is that a comparison based on one protocol can produce a model-selection decision that does not survive another legitimate protocol. The stability of that decision should therefore be measured rather than assumed.

We introduced the Protocol Sensitivity Score and a reference-based framework for this purpose. PSS measures displacement of the complete ranking, while winner swap records the decision-level consequence. The reference framework compares the between-protocol displacement with displacement under selected, fully refitted intraprotocol perturbations that preserve the scored targets. These measures complement conventional performance estimates and significance tests; they do not replace them.

Within the validated scope---daily PM10, two temporal protocols, up to nine models, and two independent monitoring networks---the results support reporting ranking stability alongside claims of model superiority. A small set of legitimate intraprotocol perturbations can provide an empirical reference for interpreting how sensitive the selected model is to evaluation choices. Whether the same framework produces similar findings outside sequential prediction remains a question for future work (Section~\ref{sec:future}).

A predictive-model comparison produces more than an error estimate. It also supports a decision. In the setting studied here, that decision changes frequently across the two evaluation protocols and at a substantially higher rate than under the selected intraprotocol perturbations.

\section{Reproducibility statement}
\label{sec:repro}

The public inputs are the frozen Heisig/EEA snapshot \citep{heisig2024} and EPA AirData \citep{epa_airdata}. The analysis code is available at \url{https://github.com/rafa-rodriguess/protocol-induced-uncertainty-predictive-model-evaluation_pub}. The notebook \texttt{orchestrator.ipynb} runs the eligibility filters, fits the protocol configurations, executes the statistical analyses, calculates PSS, and regenerates the manuscript figures and tables.

All reported numerical values are obtained from a versioned consolidated results file rather than entered manually. The study follows the machine-learning reproducibility checklist of \citet{pineau2021}. Supplementary Section~SM-01 provides the item-level compliance table, including fixed hyperparameters, run counts, and recovered wall-clock runtimes. Supplementary Sections~SM-01 and SM-02 document equivalent stage runners, dependency versions, random seeds, hardware, and the protocol-freeze record.

\section*{Declaration of generative AI and AI-assisted technologies}
During preparation of this manuscript, the authors used Claude models (Anthropic) and ChatGPT (OpenAI) to translate text and improve wording and readability. These tools were not used to design the study, formulate the hypotheses, execute the analyses, generate the reported results, or make the scientific interpretations. The authors reviewed and edited all AI-assisted text and take full responsibility for the content of the manuscript.


\bibliography{ref}
\bibliographystyle{tmlr}

\end{document}


\maketitle

\noindent\textbf{Version.} SM-01 v1.1, coordinated with the revised main manuscript, which now contains the general PSS formulation, attribution rule, audit procedure, and essential inferential settings.\\
\textbf{Contents.} Every section begins on a new page. Figure and table numbers use the SM- prefix.

\tableofcontents
\newpage

\section{SM-01 Reproducibility Workflow}
\label{sm:repro}

This supplement documents the reproducibility workflow for the main manuscript. The study follows the machine-learning reproducibility checklist proposed by \citet{pineau2021} (NeurIPS Machine Learning Reproducibility Checklist v2.0). The item-by-item mapping is the versioned design document \texttt{notes/phase5/S3\_NeurIPS\_checklist\_draft.md}; Table~\ref{tab:sm-pineau} summarises compliance for this submission.

\subsection{Prior specification}

Design, eligibility criteria, statistical plan, and decision thresholds were fixed and versioned before execution, in a dated document, without registration in an external repository. Being an internal mechanism, it does not replace independent verification, and it is not what claim credibility rests on alone. Protocol-freeze chronology is detailed in Supplementary Section~\ref{sm:freeze}.

\subsection{Confirmation under a frozen protocol}

The confirmation procedure is documented by its execution order: the full protocol specification was frozen and dated before confirmation material was selected and before any metric was computed on it, and execution occurred once, without adjustment. Freeze, selection, and execution timestamps appear in versioned artefacts and can be confronted with each other (Supplementary Section~\ref{sm:freeze}).

\subsection{Robustness checks and alternative explanations}

The study records the robustness checks and alternative explanations examined during the analysis. Each was associated with an executable test and an interpretation rule specified before the corresponding execution; the main manuscript reports their outcomes. Two checks led to revisions of manuscript conclusions, and both are documented with the preceding values preserved in the versioned artefacts. Supplementary Section~\ref{sm:freeze} enumerates each check with its date, criterion, and outcome, and distinguishes decisions fixed before analysis from those that emerged during the work.

\subsection{Data, licences, and ethics}

Inputs are public and citable: the Heisig/EEA snapshot under Zenodo DOI \citep{heisig2024} and the EPA AirData network in the public domain \citep{epa_airdata,epa_aqs}. Using a frozen archive rather than live service access is what makes reproduction byte-for-byte verifiable.

The Heisig/EEA snapshot is reused with attribution under the snapshot and EEA service terms \citep{heisig2024}; EPA AirData are public domain \citep{epa_airdata}. As environmental monitoring data, the study involves no human subjects and does not require ethics approval.

\subsection{Code}

\begin{sloppypar}
A public code repository contains the analysis orchestrator (\texttt{orchestrator.ipynb}), the production package \texttt{pss\_audit} (data ingest, protocols, PSS, statistical battery, artefact registry), stage runners under \texttt{scripts/}, \texttt{requirements.txt}, and an exact dependency lockfile \texttt{requirements.lock.txt}. The orchestrator is the recommended single entry point for executing the full confirmatory pipeline and regenerating manuscript figures and tables. When \texttt{output/metadata/study\_results\_master.json} is already present, figures and tables can be regenerated without refitting models. A DOI deposit may be added upon acceptance.
\end{sloppypar}

\subsection{Numerical artefacts}

All values reported in the main manuscript derive from a single consolidated results file, versioned with the code, recording each pipeline macro-stage outcome and the decision associated with each pre-specified gate. Tables and figures are generated from it, without manual transcription.

\subsection{Fixed hyperparameters and run counts}

Hyperparameters were \textbf{fixed a priori}; there was no hyperparameter search. Production knobs (distinct from smoke-test configurations) are recorded in \texttt{pss\_audit/config.py}: \texttt{RANDOM\_SEED}=42; horizons \(h=1,\ldots,7\); evaluation window of two years; static train fraction 0.75; rolling update \texttt{MS}; XGBoost with 100 estimators and maximum depth 4; SARIMA order \((1,0,1)\) with weekly seasonal \((1,0,0,7)\); maximum lag 14; MCS $\alpha=0.10$; TOST margin $\delta=0.10$ for the three-model arm. The confirmatory protocol was executed once per station on each panel arm; the frozen-protocol confirmation set was likewise run once after freeze. Uncertainty uses block bootstrap with $B=2000$. Model-fit workload scales as approximately $K|\mathcal{U}||\mathcal{M}|$ fit streams across protocol configurations, stations, and models, as stated in the main manuscript.

\subsection{Environment and runtime}

Analyses were executed on an Apple MacBook Air with an Apple M4 processor (10 cores: 4 performance and 6 efficiency) and 24\,GB of unified memory, running macOS 26.5.1. Language, library, and random-seed versions are recorded in the repository (\texttt{requirements.txt}, \texttt{requirements.lock.txt}, and stage metadata under \texttt{output/metadata/}).

Wall-clock runtimes were recovered from existing stage metadata without re-executing the confirmatory pipeline (\texttt{output/metadata/runtime\_summary.json}). Recorded median per-station protocol runtime is approximately 15.2\,s on the European fitting path and 17.4\,s on the US EPA panel ($N=365$), corresponding to about 1.85\,h wall time for that EPA protocol pass. Energy or carbon cost was not instrumented and is marked not applicable in Table~\ref{tab:sm-pineau}.

\subsection{Pineau checklist compliance}

Table~\ref{tab:sm-pineau} confirms application of the \citet{pineau2021} checklist. Status codes: Yes (satisfied and documented), Partial (present but incomplete on the manuscript/repo surface), NA (not applicable). No checklist item remains unanswered as No for this submission; the two Partial rows are documentation depth (expanded complexity/runtime detail and unmeasured energy), not missing protocol elements.

\begin{table}[htbp]
\centering
\caption{Machine Learning Reproducibility Checklist v2.0 \citep{pineau2021}: compliance for this study. Full comments and file pointers: \texttt{notes/phase5/S3\_NeurIPS\_checklist\_draft.md}.}
\label{tab:sm-pineau}
\small
\begin{tabular}{@{}p{2.2cm}p{6.0cm}cc@{}}
\toprule
Domain & Checklist item & Status & In SM/main \\
\midrule
Models & Clear description of setting / algorithm / model & Yes & Methods \\
Models & Clear explanation of assumptions & Yes & Methods, limits \\
Models & Complexity analysis (time / space / sample) & Partial & Main + runtime note \\
Theory & Clear statement of claim & Yes & Intro / RQs \\
Theory & Complete proof & NA & Empirical paper \\
Data & Relevant statistics (sample sizes) & Yes & Data, tables \\
Data & Train / validation / test split details & Yes & Methods, SM-02 \\
Data & Exclusions and preprocessing & Yes & Data, SM-09 \\
Data & Downloadable dataset link & Yes & Heisig DOI, EPA \\
Data & New data-collection process & NA & Public archives \\
Code & Dependency specification & Yes & \texttt{requirements*.txt} \\
Code & Training code & Yes & \texttt{pss\_audit}, orch.\ \\
Code & Evaluation code & Yes & PSS / stats battery \\
Code & Pre-trained model release & NA & Full refit policy \\
Code & README + results table + run command & Yes & Public README \\
Results & Hyperparameters, selection, finals & Yes & Fixed; no search \\
Results & Exact number of train / eval runs & Yes & Once + $B=2000$ \\
Results & Definition of reported measures & Yes & PSS, swap, TOST \\
Results & Central tendency and variation & Yes & Mean + block CI \\
Results & Average runtime or energy cost & Partial & Runtime yes; energy NA \\
Results & Computing infrastructure & Yes & This section \\
\bottomrule
\end{tabular}
\end{table}

\clearpage

\section{SM-02 Protocol Freeze and Pre-commitment}
\label{sm:freeze}

\subsection{Genealogy of decisions and adversarial checks}

Chronological enumeration of each design decision, when it was fixed, and its motivation; and, for each alternative explanation tested, the reading rule, execution date, and outcome. Explicitly records which elements were fixed before any analysis---eligibility criteria, index viability threshold, equivalence margin---and which emerged during the work, including the attribution hypothesis, formulated after observing the first placebos and tested under a rule fixed before its execution. Includes the two conclusion revisions in the process: the operational definition of hard reversal, changed after finding that the confidence-set criterion was inert with three models; and discrepancy attribution, revised after strict pairing of the evaluation sample.

Without registration in an external repository, the claim that criteria were fixed before analysis is not third-party verifiable. The mechanism that replaces it is: the full protocol specification---models, validation protocols, horizons, eligibility, PSS and winner-swap definitions, equivalence margin, statistical plan, and block definition---was frozen and dated \textbf{before} confirmation material was selected, and selection used a structural criterion with no metric-based filtering. No quantity had been computed on those stations at selection time. The protocol was then run once, without adjustment. This ordering prevents adaptation of the frozen procedure to results from the subsequently selected confirmation material.

\subsection{Freeze timestamp for confirmation material}

Frozen at: 2026-07-20T12:19:33Z; blindness: BLIND (TI exception ES/ES1968A). Repository history audit confirms that no metric had been computed on the 271 reserved background stations before freeze, with one declared exception: one traffic/industrial station---the same used in the instrument check (Supplementary Section~\ref{sm:instrument})---had a directional reversal metric computed two days earlier.

\subsection{Pre-specified criteria and outcomes}

Table~\ref{tab:sm-gates} collects all pre-analysis criteria, obtained values, and decisions.

\begin{table}[htbp]
\centering
\caption{Pre-specified criteria and outcomes.}
\label{tab:sm-gates}
\small
\begin{tabular}{@{}p{4.2cm}p{3.2cm}p{2.4cm}p{2.4cm}@{}}
\toprule
Criterion & Threshold & Obtained & Decision \\
\midrule
EEA sample floor & 250 & 1216 & pass \\
External sample floor & 50 & 365 & pass \\
Instrument check ($\mathrm{PSS}\ge 1.0$) & 1.000 & 1.143 & pass \\
Equivalence margin $\delta(M=3)$ & 0.100 & 0.100 & locked \\
Equivalence margin $\delta(M=9)=4/[M(M-1)H]$ & 0.008 & 0.008 & locked \\
Index viability (adj.\ $R^2$) & 0.150 & 0.105 & supplementary \\
Exceedance event-count rule & $n$ filter & supplementary & supplementary \\
Placebo / mechanism / recency reading & pre-specified labels & confounder discarded & confounder discarded \\
Held-out panel reading & confirm if CI-compatible & confirm & confirm \\
\bottomrule
\end{tabular}
\end{table}

\subsection{Complete execution sequence}

\begin{figure}[htbp]
\centering
\includegraphics[width=\linewidth]{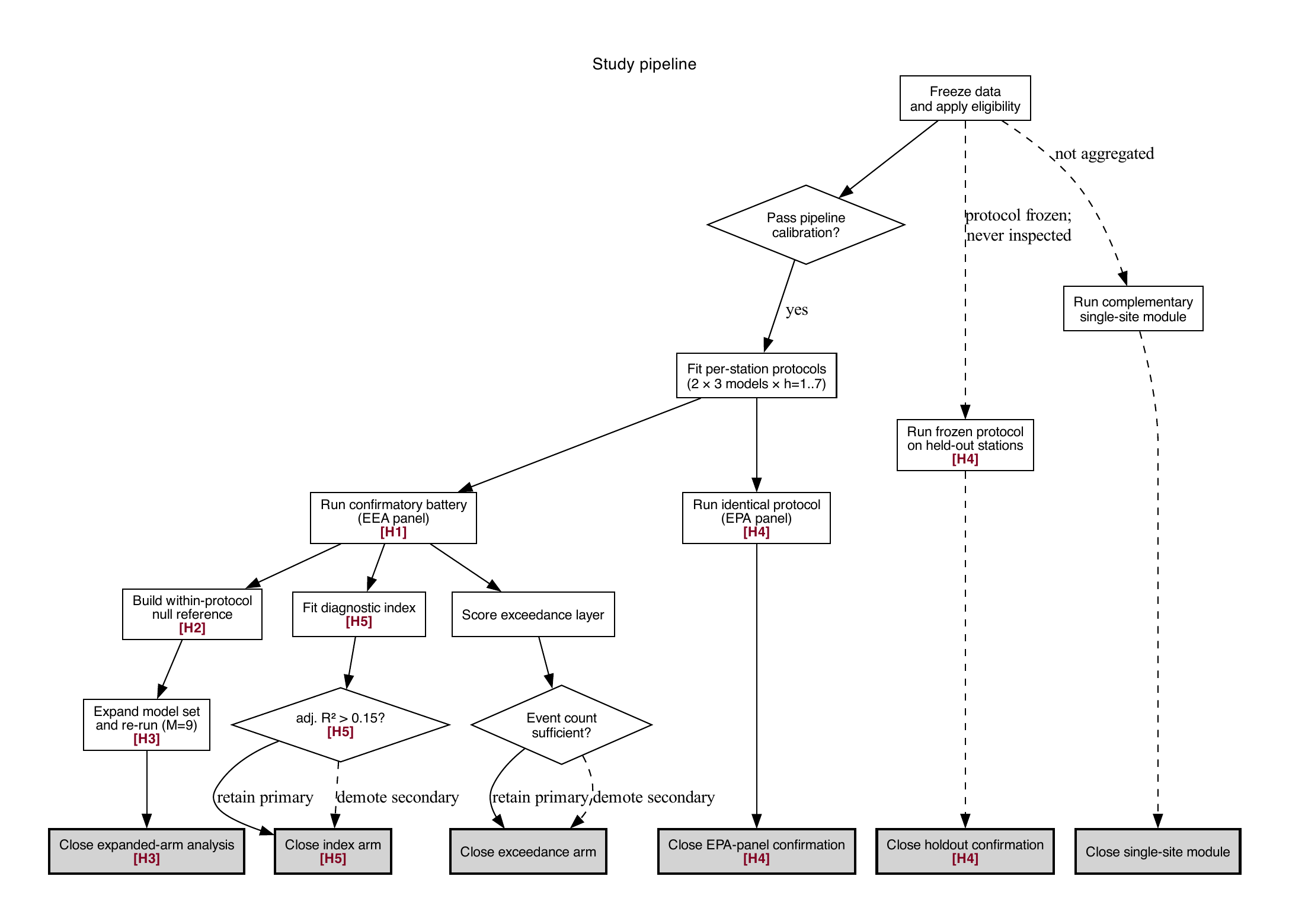}
\caption{Macro execution sequence with hypothesis labels \texttt{[Hn]}, explicit branch terminals, and parallel European and US arms.}
\label{fig:sm-pipeline}
\end{figure}

Figure~\ref{fig:sm-pipeline} records the full operational sequence, including secondary branches and the pre-specified gates that determine whether they remain primary or supplementary. The main manuscript retains only the scientific sequence needed to understand H1--H4.

\clearpage

\section{SM-03 Statistical Details}
\label{sm:stats}

\subsection{Choice of concordance measure}

We use Kendall's $\tau$ because the scientific object is pairwise ordinal concordance between two \emph{complete short rankings}, not association between continuous scores: each discordant pair is an operational inversion of model order \citep{kendall1938,arndt1999,puth2015}. The transform $\mathrm{PSS}=1-\tau$ merely reorients concordance into a displacement scale whose zero is ranking identity. Horizon averaging is required by the multi-step decision object ($h=1,\ldots,7$) and supplies the discrete resolution that anchors $\delta$; it also avoids selecting a single horizon after the fact. Spearman's $\rho$ would treat ranks as scores; for $M=3$ complete rankings without ties the two coefficients are nearly monotone transforms of each other, so panel-level qualitative conclusions would be expected to agree, but Kendall remains the natural pairwise reading. Top-1 disagreement is not omitted: it is the secondary measure (hard reversal / winner swap) in the main manuscript. Top-weighted indefinite-list similarities such as rank-biased overlap \citep{webber2010} target long, incomplete rankings and introduce an untuned persistence parameter; they are a poor match to conjoint three-model rankings and are not used here.

\subsection{MCS-exclusion severity criterion}

An alternative severity criterion anchored in inference---requiring the static-split winner to lie \textbf{outside} the Model Confidence Set \citep{hansen2011} estimated under rolling-origin ($\alpha=0.10$) on at least four of seven horizons---was computed in parallel and proved practically inert in this design: it holds at 0.24\% of European stations with three models and 1.4\% with nine. The reason is the width of confidence sets under rolling-origin; the result is reported here as a methodological finding, not as a measure of the phenomenon. For transparency, a PSS variant restricted to MCS-distinguishable pairs was listed among planned outputs but never computed---the corresponding column replicates raw PSS---and is therefore unused in any result.

MCS-exclusion prevalence from $\ge 1/7$ to $\ge 7/7$ horizons is 16.9\%, 3.3\%, 0.24\%, 0.24\%, 0\%, 0\%, 0\% on the European panel, documenting that criterion's inertia.

\subsection{Confirmatory-battery detail}

The main manuscript reports the essential inferential settings ($B=2000$, country/state spatial blocks, the station-level FDR family, the grid-anchored TOST margin, and one-grid-step power). For auditability, the underlying European battery contains 8925 pairwise Diebold--Mariano tests, aggregated into one multiplicity-controlled decision per station. The exact station decision rule, block memberships, resampling seeds, and TOST outputs are stored in the versioned numerical artefacts.

Exploratory analyses comprise Wilcoxon and sign tests for $\mathrm{PSS}>0$; Moran's $I$ (nearest neighbours, $k=5$) for spatial dependence diagnosis; and robustness checks on the hard-reversal threshold, seasonality, imputation sensitivity, exclusion of 2020--2021, and alternative spatial blocks. These analyses are descriptive and support no main claim.

\clearpage

\section{SM-04 Instrument Validation}
\label{sm:instrument}

Before interpreting the full panel, we verify that the instrument can detect the phenomenon it claims to measure. At a site where ranking displacement is expected, the pre-specified criterion of hard reversal or $\mathrm{PSS}\ge 1.0$ was met (station \textbf{ES/ES1968A}: $\mathrm{PSS}=1.14$ with reversal; Table~\ref{tab:sm-gate}). This is an instrument sanity check, not a validity condition for the claims---no main result depends on that station---and the criterion is conceptual (direction and magnitude of instability), not pointwise numerical reproduction of an external study.

\begin{table}[htbp]
\centering
\caption{Instrument check before the panel.}
\label{tab:sm-gate}
\begin{tabular}{@{}llll@{}}
\toprule
Station & Pre-specified criterion & PSS & Reversal \\
\midrule
ES/ES1968A & hard reversal or $\mathrm{PSS}\ge 1.0$ & 1.143 & yes \\
\bottomrule
\end{tabular}
\end{table}

\clearpage

\section{SM-05 Robustness and Null Diagnostics}
\label{sm:robust}

\subsection{Null reference and robustness checks}

The main manuscript reports the paired/unpaired headline comparison because target identity is constitutive of the estimand. This section archives the full-refit distributions, block intervals, station-level paired differences, target-pairing verification, and the preliminary unpaired outputs for both the three- and nine-model arms.

Additional diagnostics include the prevalence curve under the MCS-exclusion criterion and its placebo counterpart; PSS by horizon with intervals; the ranking-transition table between protocols; intervals under alternative spatial-block definitions; a static-split variant with training contiguous to evaluation; exclusion of 2020--2021; and imputation-rule sensitivity. These outputs document that the attribution result is not created by one block definition, training-recency gap, pandemic interval, or gap-treatment convention.

\subsection{Additional controls (three-model decomposition)}

Additional controls accompanying the main-text decomposition: no-recency-gap (V8) 0.773 [0.638; 0.883]; preliminary unpaired placebo 0.721; paired $\Delta$ (cross $-$ rolling) 0.571 [0.425; 0.674]; fraction above placebo median 0.96; above placebo p95 0.75; pairing fraction 1.0.

\clearpage

\section{SM-06 Diagnostic Index (H5)}
\label{sm:index}

\subsection{Full specification}

The diagnostic index asks whether a station's expected PSS can be predicted from observable series features, with volatility as the primary predictor defined a priori---under the conjecture, recorded before analysis, of an \textbf{inverse} relation between volatility and protocol sensitivity---and remaining predictors as exploratory. Validation is spatially blocked so that stations from the same block do not appear simultaneously in train and test.

The index was submitted to a viability criterion fixed before analysis: adj.\ $R^2>0.15$ to appear as a main-text contribution, otherwise supplementary. The fit obtained was adj.\ $R^2=0.105$ ($R^2=0.118$), below threshold, so the index is reported only here. Two consequences follow and were respected: the index is not scored on the EPA panel, and the cross-continental test evaluates transfer of the \emph{phenomenon}, not of the predictor.

The fitted coefficients are consistent with retaining the index as a supplementary result. The primary predictor's coefficient, the volatility measure, is positive ($\approx 0.137$), whereas the conjecture recorded before analysis predicted an inverse relation; that is, the only predictor with a pre-specified direction points the opposite way. We report the sign without proposing a post hoc mechanism, because the overall fit does not support interpretation of an isolated coefficient.

Full specification, estimated coefficients, adj.\ $R^2=0.105$ against the 0.15 threshold, and performance by spatial block are summarised in Figure~\ref{fig:sm-index}. The result is therefore retained in the supplementary material under the pre-specified viability rule.

\begin{figure}[htbp]
\centering
\includegraphics[width=0.80\linewidth]{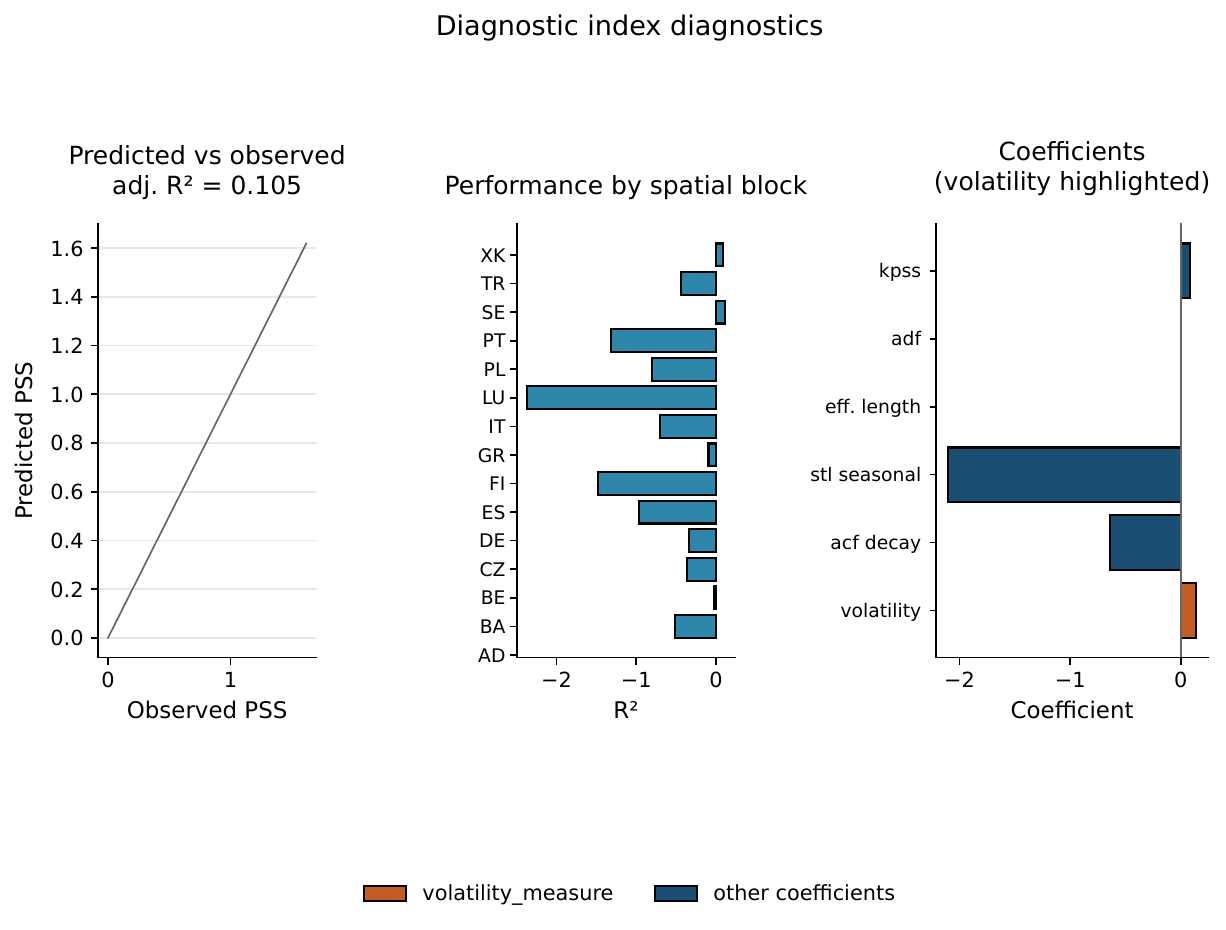}
\caption{Diagnostic-index fit diagnostics (adj.\ $R^2=0.105$ below the pre-specified viability gate of 0.15; reported in supplementary material by rule).}
\label{fig:sm-index}
\end{figure}

\clearpage

\section{SM-07 Exceedance Layer}
\label{sm:exceedance}

Exceedance-day evaluation against the regulatory PM10 threshold follows a recurring recommendation in the operational literature \citep{vitali2023,bertrand2023}, with the caveat, recorded a priori, that background stations tend to produce few events and thus low power. In this round only event counts were obtained; contingency metrics (POD, FAR, CSI) were not computed. A pre-specified event-count rule promotes the layer to main text only when counts suffice; here the rule places the layer in the supplementary material.

Against the regulatory threshold of 50~$\mu$g/m$^3$, requiring at least 10 events per station, 399 stations qualify, with 62{,}017 exceedance days in total. The restriction is not of aggregate event volume but of per-station power in the background stratum---the caveat recorded a priori---and therefore the layer is not promoted to main text. Contingency metrics remain for later work.

\begin{table}[htbp]
\centering
\caption{Exceedance-layer tier (Supplementary Section SM-07).}
\label{tab:sm-exceed}
\resizebox{\textwidth}{!}{%
\begin{tabular}{@{}cccccc@{}}
\toprule
Threshold ($\mu$g\,m$^{-3}$) & Min events & Stations $\ge$ min & Exceedance days & Tier & POD/FAR/CSI \\
\midrule
50 & 10 & 399 & 62017 & SUPPLEMENTARY & not computed \\
\bottomrule
\end{tabular}%
}
\end{table}

\clearpage

\section{SM-08 Single-site Mechanism Module}
\label{sm:singlesite}

A parallel module, restricted to one station and analysed without aggregation into the panel, serves two functions: re-execute on the available series version the single-station result that motivated the panel-scale study, and apply the inferential layer that the manuscript declares unused---Diebold--Mariano with HLN correction, block-bootstrap intervals over origins, and per-horizon confidence sets. The response analysed is skill relative to persistence.

\subsection{Mechanism-module station (Elche)}

The complementary module operates on the urban-background station at Elche (Alicante), the site of the case reported by \citet{garciacrespi2026}. That manuscript uses the regional Comunitat Valenciana network over 2017--2024; we use the EEA version of the same station for the same interval. The two compilations match closely on distributional moments---mean 20.0 vs 20.1~$\mu$g\,m$^{-3}$, median 18.0 in both, maximum 220.0 vs 218.3---and diverge in completeness, with 84.3\% valid days in the EEA version. With the completeness regime of the original study, the station would not meet our panel eligibility criterion; the module is therefore analysed separately and never aggregated into panel analyses.

\subsection{Where the origin station falls in the panel}

Applying the panel protocol to the Elche station without adaptation---same train fraction, evaluation window, origins, models, and horizons---its PSS is 0.286, against a median of 0.857 and IQR 0.571 to 1.048 among the 425 European stations. The station lies at percentile 8.9: \textbf{it is among the 10\% least sensitive in the panel}, outside the IQR on the lower edge, and shows no winner swap on any of the seven horizons. Its per-horizon profile is also atypical: discrepancy is null at $h=1$ and $h=2$ and concentrates at intermediate horizons, whereas in the panel the maximum occurs precisely at $h=1$.

This admits two readings, and both should be stated. The first reinforces the phenomenon's reach: the reported case describes a reversal at a site that, by our measure, is among the most stable on the continent---if rankings moved there, the typical station, three times more sensitive, offers much more fertile ground. In that sense the finding was conservative. The second is a scope caveat: under our protocol and on the series version we have, we do not observe at that station the ranking swap reported there, which is coherent with the re-execution result below and reinforces that the two measurements are not interchangeable.

There is, however, a coincidence that supports the generality of the panel attribution. At the origin station, the fixed-point placebo produces 0.095 against 0.286 for the between-protocol contrast---a ratio of about three to one. On the European panel the ratio is 0.230 to 0.801, or about three and a half to one. Absolute instability magnitude varies enormously across sites; the \textbf{proportion} of protocol effect to internal noise does not.

\subsection{Re-execution under completeness regimes}

Re-execution under EEA completeness does not reproduce the reported reversal (XGBoost skill at $h=1$: $+0.128$ vs $-0.192$); under a restrictive quality filter approximating the original study's completeness, XGBoost's short-horizon failure reappears ($-0.148$) but SARIMA dominance does not. A factorial grid of 30 refit-calendar phases shows that phase accounts for 88.5\% of $h=1$ skill variance among the perturbations tested.

\subsection{Inferential layer}

None of the three $h=1$ Diebold--Mariano comparisons is significant---SARIMA vs persistence ($p=0.288$), XGBoost vs persistence ($p=0.144$), SARIMA vs XGBoost ($p=0.283$). SARIMA skill at $h=1$ of 0.099 has confidence interval $-0.113$ to 0.222, indistinguishable from zero.

\clearpage

\section{SM-09 Eligibility Census and Per-station Archive}
\label{sm:eligibility}

\subsection{Per-station results}

PSS, hard-reversal flag, and Diebold--Mariano decision for each station of both panels are provided in a supplementary data file accompanying this document.

\subsection{Detailed eligibility}

Census by country and by state, and characterisation of the inventoried EEA/airbase set excluded from the confirmatory panel, appear in the versioned design artefacts. The EEA/airbase set (2615 stations, 1172 eligible) remains entirely outside confirmatory inference for provenance reasons stated in the main manuscript.

\section{SM-10 Aggregation Stability Diagnostics}
\label{sm:agg}

The main manuscript's aggregation-stability section reports Curves~A and~B (aggregate one protocol; compare to the other protocol's reference). This supplement archives the complementary Curve~C---Borda aggregation of $K$ static runs and $K$ rolling runs, then hard-reversal between the two aggregated winners---and the discard log for grid cells that failed exact scored-pair pairing or the static train-fraction gate $[0.65,0.85]$ ($\AggStabNDisc$ discarded cells across both panels; all retained cells have pairing fraction $1.0$).

Curve~C uses the same seed ($42$), subset count ($100$), and hard-reversal rule ($\ge 4/7$ horizons) as the main-text curves. Table~\ref{tab:aggc} reports the median [IQR] winner-swap rates. As on Curves~A and~B, aggregation moves the rate only marginally and leaves a large residual between-protocol gap. The nine-model arm was not repeated: the main manuscript already shows that expanding the candidate set amplifies selection instability, and the $M=3$ aggregation curve is the fundamental calibration.

\begin{table}[htbp]
\centering
\caption{Curve~C winner-swap rate (\%) under joint Borda aggregation: median [IQR] over 100 subsets.}
\label{tab:aggc}
\begin{tabular}{@{}lccc@{}}
\toprule
Panel & $K$ & Median [\%] & IQR [\%] \\
\midrule
EEA & 1 & 37.6 & [35.5; 40.0] \\
EEA & 3 & 35.5 & [33.9; 37.2] \\
EEA & 5 & 34.8 & [33.4; 35.5] \\
EEA & 7 & 34.6 & [33.4; 35.3] \\
EEA & 10 & 35.5 & [34.6; 36.2] \\
EEA & 13 & 34.4 & [33.9; 34.6] \\
EPA & 1 & 32.3 & [31.0; 36.4] \\
EPA & 3 & 30.4 & [29.6; 31.5] \\
EPA & 5 & 29.9 & [29.0; 30.7] \\
EPA & 7 & 29.6 & [28.8; 30.4] \\
EPA & 10 & 30.1 & [29.2; 31.2] \\
EPA & 13 & 29.3 & [29.0; 29.6] \\
\bottomrule
\end{tabular}
\end{table}

\bibliography{ref}
\bibliographystyle{tmlr}